\documentclass{article} 
\usepackage{acl}
\usepackage{times}
\usepackage{latexsym}

\usepackage{amsmath,amsfonts,bm}

\def\eqref#1{equation~\ref{#1}}

\def\1{\bm{1}}

\DeclareMathAlphabet{\mathsfit}{\encodingdefault}{\sfdefault}{m}{sl}
\SetMathAlphabet{\mathsfit}{bold}{\encodingdefault}{\sfdefault}{bx}{n}

\usepackage{xcolor}
\usepackage{amsthm}
\usepackage{amsfonts}
\usepackage{bm}
\usepackage{amsmath}
\usepackage{amssymb}
\usepackage{wrapfig}
\usepackage{float}
\usepackage{wasysym}
\usepackage{mathtools}
\usepackage{enumitem}
\usepackage{graphicx}
\usepackage{cases}
\usepackage{microtype}

\setlist{nosep}
\setlist[itemize]{leftmargin=*}

\definecolor{lightindigo}{RGB}{242, 244, 255}
\definecolor{lightanthracite}{RGB}{252, 252, 255}
\definecolor{blueanthracite}{RGB}{61,71,94}
\definecolor{indigo}{RGB}{63, 81, 181}
\definecolor{red}{RGB}{210, 40, 95} 
\definecolor{pink}{RGB}{236, 64, 122}
\definecolor{green}{RGB}{46, 182, 125}
\definecolor{blue}{RGB}{3, 81, 154}
\definecolor{yellow}{RGB}{236, 178, 46}
\definecolor{anthracite}{RGB}{61, 61, 81}
\definecolor{gold}{RGB}{182, 131, 76}
\definecolor{lightgrey}{RGB}{128, 128, 128}
\definecolor{purple}{RGB}{156, 39, 176}

\renewcommand{\vec}[1]{\bm{#1}}

\newcommand{\x}{\vec{x}}

\newcommand{\z}{\bm{z}}

\newcommand{\deltaz}{\Delta \mathbf{z}}
\newcommand{\deltac}{\Delta \mathbf{c}}
\newcommand{\hatdeltac}{\Delta \hat{\mathbf{c}}}
\newcommand{\hatdeltaz}{\Delta \hat{\mathbf{z}}}

\usepackage{hyperref}
\usepackage{url}
\usepackage{booktabs}
\usepackage{todonotes}
\usepackage{xcolor}
\usepackage{wrapfig}
\usepackage{tikz-cd}
\usepackage{enumitem}
\usepackage{multirow}
\usepackage{inconsolata}
\usepackage{subcaption}
\usepackage{fvextra}
\DefineVerbatimEnvironment{Resume}{Verbatim}{
  breaklines,
  breaksymbolleft={},
  fontsize=\small,
  commandchars=\\\{\}
}

\usepackage{listings}
\usepackage[skins,breakable,raster]{tcolorbox}
\tcbuselibrary{listings}

\tcbset{
  listing engine   = listings,
  listing options  = {
    breaklines=true,
    basicstyle=\ttfamily\footnotesize,
    columns=flexible,
    breakatwhitespace=false
  },
  boxrule=0.35pt, sharp corners,
  left=0.8ex, right=0.8ex, top=0.5ex, bottom=0.5ex,
  negbox/.style  = {colframe=red!45,   colback=red!3},
  nonebox/.style = {colframe=black!30, colback=black!3},
  posbox/.style  = {colframe=blue!45, colback=blue!3},
}

\newtcolorbox{qbox}[1][]{
  colframe=black!30,
  colback=white,
  title={#1},
}

\newtcolorbox{respbox}[1][]{
  #1
}

\usepackage{cleveref}
\crefname{section}{\S}{\S\S}
\crefname{subsection}{\S}{\S\S}
\crefname{subsubsection}{\S}{\S\S}
\crefname{figure}{Fig.}{Figs.}
\crefname{prop}{Prop.}{Props.}
\crefname{appendix}{Appx.}{Appxs.}
\crefname{algorithm}{Alg.}{Algs.}
\crefname{theorem}{Thm.}{Thms.}
\crefname{definition}{Defn.}{Defns.}
\crefname{cor}{Corollary}{Corollaries}
\crefname{lem}{Lem.}{Lems.}
\crefname{table}{Tab.}{Tabs.}
\crefname{assum}{Assum.}{Assums.}
\crefname{example}{Ex.}{Exs.}

\newtheorem{definition}{Definition}

\definecolor{pastelbrown}{HTML}{bb9e8f}
\definecolor{pastelgreen}{HTML}{8fbb9e}
\definecolor{pastelpurple}{HTML}{ac9fc5}

\NewDocumentCommand\todoAM{O{}m}{\todo[author=AM,color=pastelbrown,#1]{#2}}
\NewDocumentCommand\todoESL{O{}m}{\todo[author=ESL,color=pastelpurple,#1]{#2}}
\NewDocumentCommand\todoAL{O{}m}{\todo[author=AL,color=pastelgreen,#1]{#2}}

\title{From Isolation to Entanglement: When Do Interpretability Methods Identify and Disentangle Known Concepts?}

\author{Aaron Mueller$^{1}$, Andrew Lee$^2$, Shruti Joshi$^3$, Ekdeep Singh Lubana$^4$,\\ \textbf{Dhanya Sridhar}$^3$\textbf{, Patrik Reizinger}$^5$ \\
$^1$Boston University\ \ \ \ \ $^2$Harvard University\ \ \ \ \ $^3$Mila -- Quebec AI Institute\\$^4$Goodfire\ \ \ \ \ $^5$University of Tübingen\\
}

\begin{document}

\maketitle

\begin{abstract}
    A goal of interpretability is to recover disentangled representations of latent concepts (features) from the activations of neural networks. The quality of features is typically evaluated in isolation, and under implicit independence assumptions that may not hold in practice. Thus, it is unclear to what extent common featurization methods such as sparse autoencoders (SAEs) and probes \emph{disentangle} one concept from another. We propose a multi-concept evaluation setting using concepts including sentiment, domain, voice, and tense. 
    We evaluate how well featurizers produce disentangled representations of each concept, observing
    that features are typically sensitive to only one concept, but also that concepts are distributed across many features. Then, we steer these features, measuring whether each concept is independently manipulable, and whether features interact. Even in idealized settings, steering a feature often affects \emph{many} concepts, despite a near absence of interaction effects.
    These results suggest that correlational metrics are insufficient to establish steering selectivity, and that demonstrating that two features operate in separate spaces is insufficient to claim that they will be selective for one concept. These results underscore the importance of multi-concept evaluations in interpretability research.\footnote{Data and code are available at \url{https://github.com/aaronmueller/IdentifiableLanguage}.}
\end{abstract}

\section{Introduction}
Interpretability centers on understanding and controlling neural network behaviors \citep{geiger-2025-causalabstraction,mueller2025questrightmediatorsurveying}. This requires understanding the underlying causal variables and mechanisms that produce observed input-output behaviors. To precisely localize these causal variables, \emph{featurization methods}, such as sparse autoencoders (SAEs; \citealp{olshausen-1997-sparse,bricken2023monosemanticity,huben2024sparse}), have become common. These methods map from activation vectors (wherein a dimension can have many meanings) to sparser spaces where there is a more one-to-one relationship between dimensions and concepts.

The implicit assumption underlying these applications is that if we can identify features that represent distinct concepts, then we should be able to steer those concepts by manipulating their corresponding features. But does representational disentanglement guarantee independent manipulability? Current concept identification and steering studies focus on detecting and/or steering single concepts or behaviors \citep[e.g.,][]{wu2025axbench,arditi2024refusal,marks2024the}. This tells us whether the concept is represented and can be manipulated, but leaves open the question of whether the concept representation is \textbf{independent} and \textbf{disentangled} from other concepts. How often does steering one concept affect others? Independence and disentanglement act as a ceiling for our trust in steering methods to induce similar behaviors in novel contexts---i.e., to what degree we have predictive power and selective control over the model's future behaviors.

This is not a new idea: the fields of causal representation learning (CRL; \citealp{scholkopf_towards_2021}) and disentangled representation learning \citep{higgins_towards_2018,locatello_challenging_2019,locatello_disentangling_2020} have rich literatures characterizing the assumptions under which it is possible to identify the true latent causal variables for a task. However, these fields focus on learning a representation from scratch, whereas the goal of interpretability is to derive a simplified causal model of a large and complex neural network that has already been trained. Both lines of work are unified in asking: \emph{what methods and assumptions will yield causally efficacious representations?}

\begin{figure*}
    \centering
    \includegraphics[width=0.95\linewidth]{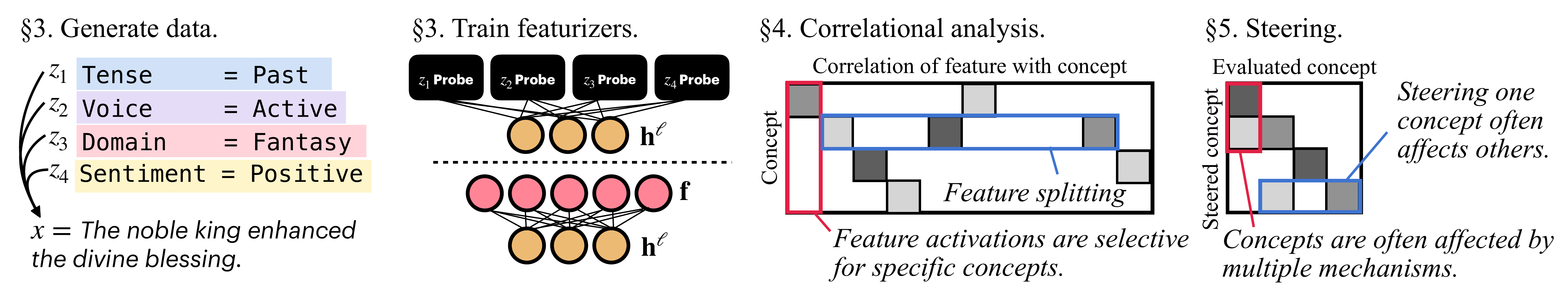}
    \caption{Overview. We first generate data where each example is labeled for multiple concepts (\S\ref{sec:experimental_setup}, App.~\ref{app:data}). Using this data, we train SAEs and probes (\S\ref{sec:experimental_setup}, App.~\ref{app:methods}). We use correlational metrics to measure how well these representations disentangle each concept, finding that they are largely disentangled (\S\ref{sec:exp_correlational}), but also that SAEs exhibit feature splitting (\S\ref{sec:exp_dcies}). However, steering the top feature for a given concept often has significant effects on \emph{many} concepts, suggesting entangled mechanisms (\S\ref{sec:exp_steer_eval}).}
    \label{fig:overview}
\end{figure*}

Our work builds upon and extends the metrics and evaluation paradigms of CRL to measure mechanistic disentanglement in a multi-concept evaluation setting. We design a dataset where each example has multiple ground-truth concept labels. We generate a natural language dataset using probabilistic context-free grammars (PCFG; \citealp{Booth1973ApplyingPM}), where each sentence is labeled with four concepts (voice, tense, sentiment, domain), and where we can control the degree of correlation between concepts in the dataset.
We use this data to evaluate common interpretability methods, including probes \citep{gurnee2023finding} and sparse autoencoders (SAEs; \citealp{olshausen-1997-sparse,huben2024sparse}), through two lenses:
(1) Do sparse features and probes achieve high scores on standard correlational disentanglement metrics from CRL (MCC, DCI-ES)? And (2) when we steer features, do they selectively manipulate their target concepts without affecting others (independence), nor each other (disjointness)?

\noindent
Our contributions and findings are:
\begin{itemize}[noitemsep]
    \item We introduce an evaluation framework and dataset to measure concept disentanglement, with adjustable correlations between ground-truth concepts (\S\ref{sec:experimental_setup}).
    \item We show that common SAE architectures and sparse probes achieve high disentanglement according to correlational metrics (\S\ref{sec:exp_identifiability}), but that this does not predict the selectivity of steering (\S\ref{sec:exp_steer_eval}). We propose metrics to quantify concept entanglement in steering (\S\ref{sec:exp_steer_heatmap}).
    \item We distinguish between feature disjointness (having non-interacting effects) and selective steerability. Current methods produce features with non-interacting effects, but steering one feature nonetheless affects multiple unrelated concepts downstream (\S\ref{sec:multisteer}).
\end{itemize}
\noindent
Our results suggest that even principled correlational metrics do not predict steering performance.
Furthermore, disjointness and independence are not mutually predictive, suggesting that current featurization objectives may be optimizing for the wrong notions of concept separation. Overall, these findings underscore the importance of multi-concept evaluations, and separating correlational from interventional evidence. 

\section{Preliminaries}\label{sec:prelim}
\textbf{Methods.} \emph{Featurization} refers to a mapping from model representations $\mathbf{h}^\ell$ to some feature space $\mathbf{f}$ that is more interpretable. We focus on two classes of featurization methods: sparse autoencoders (SAEs) and linear probes. SAEs learn an encoder $\mathcal{F}$ that maps from a representation vector $\mathbf{h}^\ell$ to a higher-dimensional feature vector $\mathbf{f}$. It also learns a decoder $\mathcal{F}^{-1}$ that reconstructs $\mathbf{h}^\ell$ from $\mathbf{f}$.\footnote{This is typically not a literal inversion. In SAEs, the decoder is learned such that the reconstruction error is minimized, but information is nonetheless lost when reconstructing $\mathbf{h}^\ell$.} The SAE is trained to reconstruct $\mathbf{h}^\ell$ accurately while also ensuring that $\mathbf{f}$ is sparse. We compare various SAE architectures; see App.~\ref{app:methods} for details. Binary linear probes learn a linear transformation from $\mathbf{h}^\ell$ to a logit; if well-trained, this logit should correlate with the probability of the concept that the probe is trained to classify.

\textbf{Metrics.} The Mean Correlation Coefficient (\textbf{MCC}; \citealp{hyvarinen_unsupervised_2016}) is a commonly used metric \citep[][\emph{i.a.}]{hyvarinen_nonlinear_2019,khemakhem_ice-beem_2020,khemakhem_variational_2020,wendong_causal_2023,von_kugelgen_self-supervised_2021,von_kugelgen_nonparametric_2023,von_kugelgen_identifiable_2024,reizinger_identifiable_2024,reizinger_jacobian-based_2023,reizinger_embrace_2023,gresele_independent_2021} that measures how well a representation recovers latent ground-truth factors, which we will refer to as ``concepts'' henceforth. Given a feature vector $\mathbf{f}$ and a set of concepts $Z$, one locates the dimension $\hat{\mathbf{f}}_i$ that has the highest correlation with concept $z_i\in Z$. The MCC gives the mean of these maximal correlations across concepts, and equals 1.0 when there exist features that identify each concept up to permutation and scaling.

\textbf{DCI-ES} \citep{eastwood_dci_2022,eastwood2023dcies} provides correlational metrics for more precisely characterizing how well a given representation disentangles concepts; we use the \textbf{D}isentanglement, \textbf{C}ompleteness, \textbf{I}nformativeness, and \textbf{E}xplicitness metrics. We first construct $R\in \mathbb{R}^{|\mathbf{f}|\times |Z|}$, where $|\mathbf{f}|$ is the dimensionality of the feature vector and $|Z|$ is the number of ground-truth concepts. Each entry $R_{i,j}$ encodes the importance of $\mathbf{f}_i$ for predicting $z_j$.\footnote{Following \citet{eastwood2018framework}, these are derived by training classifiers on $\mathbf{f}$ to predict $z_j$. We use impurity-based feature importances. Importances across features sum to 1 for a given $z_j$.} $D$ and $C$ are then defined as:
\begin{align}
    D_i &= 1+\sum_{j\in Z}R_{i,j}\log R_{i,j}\\
    C_j &= 1 + \sum_{i\in \mathbf{f}} R_{i,j}\log R_{i,j}
\end{align}
$D_i$ is inversely proportional to the entropy of the importance of feature $\mathbf{f}_i$ across concepts. $D_i$ is high (near 1) if $\mathbf{f}_i$ is only predictive of one concept $z_j$. $D_i$ is low (near 0.0) if $\mathbf{f}_i$ is equally predictive of all concepts. $C_j$ is inversely proportional to the entropy of the predictability of concept $z_j$ per feature. $C_j$ is high (near 1.0) when $z_j$ correlates strongly with only one feature's activations, and low (near 0.0) when correlated with many features.

$I$ is the negative prediction error of a classifier trained on the feature vector $\mathbf{f}$: $I_j = 1 - \mathbb{E}_{x\in \mathcal{T}}[\mathcal{L}(\mathbf{f}, z_j)]$. This measures whether concept $z_j$ can be recovered from the feature vector at all. We measure this by training linear probes on feature vectors. $E$ measures how easily concepts are recovered from the feature vector $\mathbf{f}$, proportional to the area under the loss-capacity curve. If all concepts are predictable with high accuracy using low-capacity probes, $E$ is maximized; if high-capacity probes are needed, $E$ is low.

Together, these metrics capture how easily each concept can be recovered from a feature space, whether a concept is split across features, and whether a feature is selective for only one concept. One might hypothesize that perfect scores on each would imply the ability to steer concepts in a modular way. As we show, this is not necessarily true.

\section{Experimental Setup}\label{sec:experimental_setup}

\paragraph{Data.} 
Our goal is to stress-test featurization methods by creating a dataset labeled with known concepts, but where concepts can be correlated to varying degrees. Using a probabilistic context-free grammar (PCFG),
we generate a training dataset $\mathcal{D}$ containing 382,459 sentences and test dataset $\mathcal{T}$ consisting of 1,007 sentences, where each sentence is labeled for 4 concepts $z_i \in Z$: voice, tense, sentiment, and domain. In our datasets, voice (active, passive) and tense (present, past) are binary. Sentiment (positive, neutral, negative) is multinomial and ordinal, while domain (news, science, fantasy, other) is multinomial with no inherent ordering.

To create a less idealized setting, we fix correlations between two concept values---for example, positive sentiment and the science domain. We control correlations by upsampling examples where the concept values co-occur in $\mathcal{D}$ while training featurizer $\mathcal{F}$. Under varying correlational conditions, we observe to what extent the featurizer identifies the latent ground-truth concepts. 
See App.~\ref{app:data} for further details on data generation and example sentences.

\paragraph{Models and featurizers.} 
A featurizer consists of an encoder $\mathcal{F}:\mathbb{R}^{|\mathbf{h}|}\rightarrow\mathbb{R}^{|\mathbf{f}|}$ and optionally a decoder $\mathcal{F}^{-1}:\mathbb{R}^{|\mathbf{f}|}\rightarrow\mathbb{R}^{|\mathbf{h}|}$. The encoder $\mathcal{F}$ maps hidden representation vector $\mathbf{h}^\ell$ at layer $\ell$ to features $\mathbf{f}$. 
We focus primarily on unsupervised methods such as sparse autoencoders (SAEs), due to their popularity in recent unsupervised interpretability research \citep{costa2025flathierarchicalextractingsparse,huben2024sparse,mueller2025questrightmediatorsurveying,marks2025sparse}. We formally define each SAE architecture we test in App.~\ref{app:methods}.
We compare these methods to a supervised method, linear probing.

We use two models: Pythia-70M \citep{pythia} and Gemma-2-2B \citep{gemma}. Their parameters are frozen in all experiments. We choose these because there exist publicly available SAEs trained on large natural language corpora, including the ReLU SAEs of \citet{marks2025sparse} for Pythia and JumpReLU SAEs of \citet{lieberum-etal-2024-gemma} for Gemma.

The feature vector $\mathbf{f}$ should ideally encode one concept per dimension, regardless of the correlations between concepts in the training data. Recent work has demonstrated the importance of the featurizer's inductive bias in ensuring this property, especially when deploying unsupervised featurizers \citep{hindupur2025projectingassumptionsdualitysparse,costa2025flathierarchicalextractingsparse}. We therefore compare SAEs that make varying assumptions: ReLU SAEs \citep{bricken2023monosemanticity} assume linear separability, Top-K SAEs \citep{gao2025scaling} assume angular separability, and SpADE SAEs \citep{costa2025flathierarchicalextractingsparse} make weaker assumptions that allow for more heterogeneous concept geometries. SSAEs \citep{joshi_identifiable_2025} are trained on activation \emph{differences} between pairs of inputs where concepts shift. We refer readers to App.~\ref{app:methods} for details on each SAE architecture.

\begin{figure*}[t]
    \centering
    \includegraphics[width=0.98\linewidth]{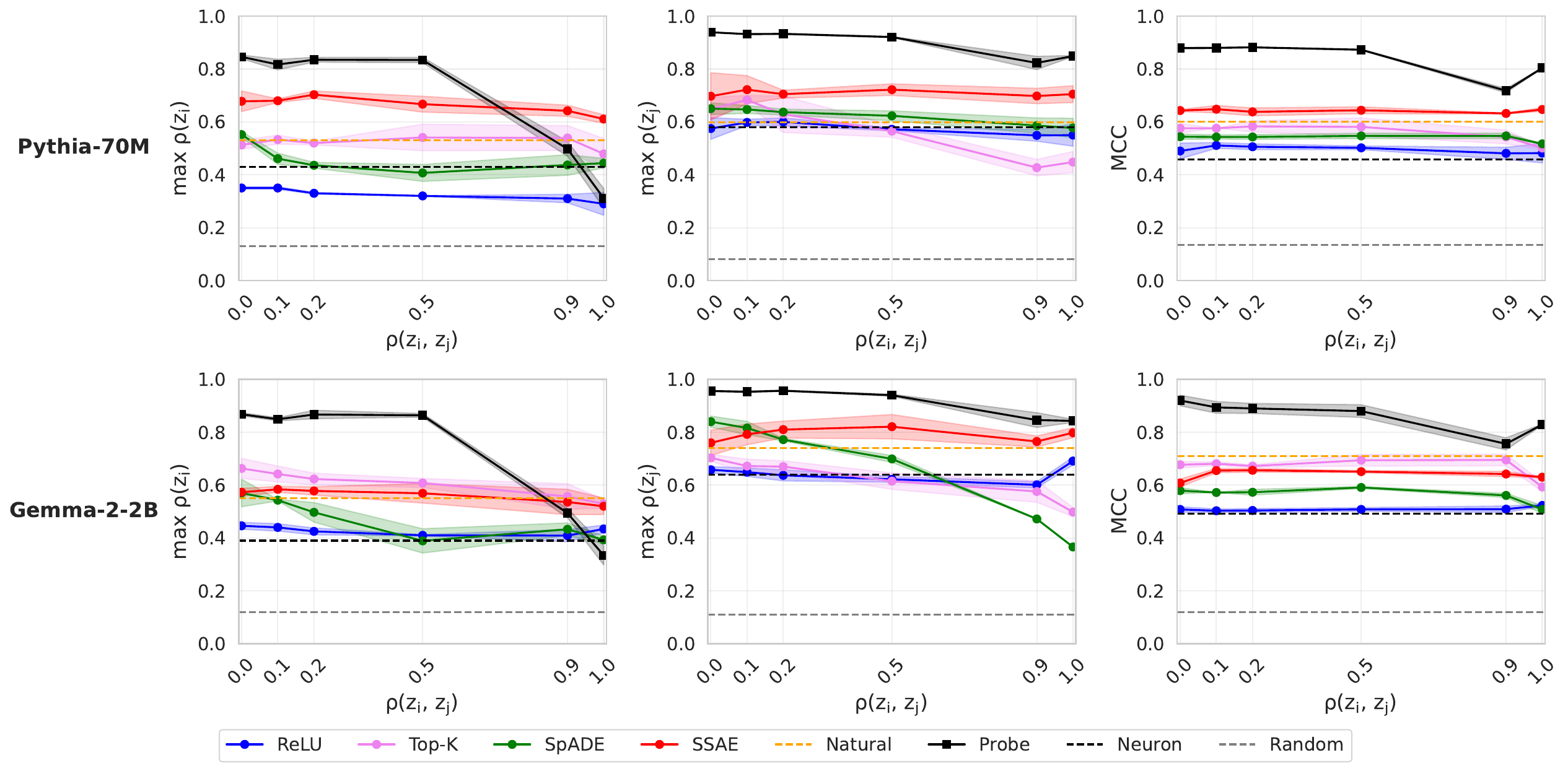}
    \caption{\textbf{Maximum correlation coefficient for domain=science (left), sentiment=positive (middle), and MCC (right) under varying correlational conditions}. Shaded regions represent 1 std.\ dev.\ across 3 training seeds. Ideal performance looks like a flat line at a high MCC. Probes (black) perform best up to cross-concept correlations of 0.9. SSAEs (red) and Top-$k$ SAEs (pink) perform best among unsupervised featurizers. SAEs trained on large-scale natural data (Natural) perform similarly to our best SAEs, but SSAEs sometimes outperform both.
    }
    \label{fig:identifiability}
\end{figure*}

\section{Features Represent Disentangled Concepts}\label{sec:exp_correlational}
\subsection{Concept identification}\label{sec:exp_identifiability}
A key desideratum of featurizers is the ability to identify the ground-truth concepts despite potential spurious correlations between them.\footnote{We cannot expect a model, supervised or unsupervised, to be able to disentangle two concepts if they are \emph{completely} correlated in the data \citep{wiedemer_compositional_2023} without making any assumptions. However, given at least a couple examples where two concepts do not covary, it is possible in theory to recover independent representations of these concepts.} To assess to what degree this property holds for popular featurizers, we design an identifiability evaluation.  Intuitively, identifiability measures whether and to what extent the learned model can recover the latent factors that generated the data (e.g., $z_i$ in Figure~\ref{fig:overview}). For formal definitions, see App.~\ref{app:sec_ident}.

\paragraph{Metrics.} To evaluate whether a featurizer recovers ground-truth concepts, we first employ the \textbf{mean correlation coefficient} (MCC; \citealp{hyvarinen_unsupervised_2016}) common in the causal representation learning literature. 
The MCC measures identifiability up to scaling and permutation (see \S\ref{sec:prelim} and App.~\ref{app:metrics}).

We correlate one feature in $\mathbf{f}$ with each concept. However, multinomial concepts may not be one-dimensional in $\mathbf{f}$ nor $\mathbf{h}^\ell$ \citep{engels2025not}. Thus, we \emph{binarize} concepts before computing MCC: given a concept $z_i$ with $|z_i|$ possible values, we create a new binary variable for each value. For example, sentiment can take a value in \{positive, neutral, negative\}. We create one binary variable per value. To compute the MCC, we first average the correlation coefficients for all values of $z_i$ before taking the macroaverage across concepts in $Z$.
A high MCC is achievable in theory only if we make the following assumption:

\textbf{Assumption:} \emph{For each value of each concept $z_i$, there exist linear transformations $T_i$ such that $z_i = T_i\mathbf{h}^\ell$, where $\mathbf{h}^\ell$ are the representations of a language model at layer $\ell$.}

To validate this assumption, we train linear probes for each binary concept value and observe whether each probe obtains high accuracy on the concept value it was trained to detect, \emph{but also} obtains random-chance accuracy on all other concepts. Our probes satisfy these criteria and thus empirically support this Assumption; see Figure~\ref{fig:probe_accuracy_heatmap_binary} (App.~\ref{app:probe_acc}).

\paragraph{Baselines and skylines.} We compare against a randomly initialized SAE (\emph{Random}); the original activation vector $\mathbf{h}^\ell$ (\emph{Neuron}, equivalent to an identity featurizer $\mathbf{f} = \mathbf{h}^\ell$); and publicly available SAEs trained on natural language data (\emph{Natural}; we use the SAEs of \citet{marks2025sparse} for Pythia-70M and GemmaScope \citep{lieberum-etal-2024-gemma} for Gemma-2-2B, respectively).

To establish a supervised skyline (\emph{Probe}), we train binary logistic regression probes for each concept value (i.e., we have separate binary probes for negative sentiment, positive sentiment, past tense, etc.). We correlate each probe's logit with ground-truth concept labels, and take the average correlation across concepts to compute the MCC.

\paragraph{Hypothesis.}
The ideal result is a high MCC that remains constant as the correlation between ground-truth concepts increases in the training data. 
Among (S)SAEs, we hypothesize that SSAEs will perform best, as sparse representations of \emph{shifts} between concepts are provably identifiable, whereas typical SAE architectures do not have this property \citep{joshi_identifiable_2025}. We expect unsupervised featurizers, such as SAEs, to perform worse than supervised featurizers, such as probes. We also expect SAEs trained on our dataset to be better able to isolate the ground-truth concepts compared to the \emph{Natural} baselines; this is because the number of varying concepts is lower, which should make these concepts easier to isolate.

\paragraph{Results.}
Figure~\ref{fig:identifiability} shows MCCs, as well as the maximum correlation coefficients for Pythia-70M and Gemma-2-2B for the domain and sentiment concepts as they become more correlated in the training dataset.
Probes significantly outperform SAEs, as expected (up to correlations of 0.9). The margin between probes and SAEs is generally substantial; thus, if one knows \emph{a priori} what concepts one wishes to find, one should use supervised methods. This agrees with recommendations from \citet{wu2025axbench} and \citet{mueller2025mib}.

\begin{figure*}[t]
    \centering
    \includegraphics[width=0.98\linewidth]{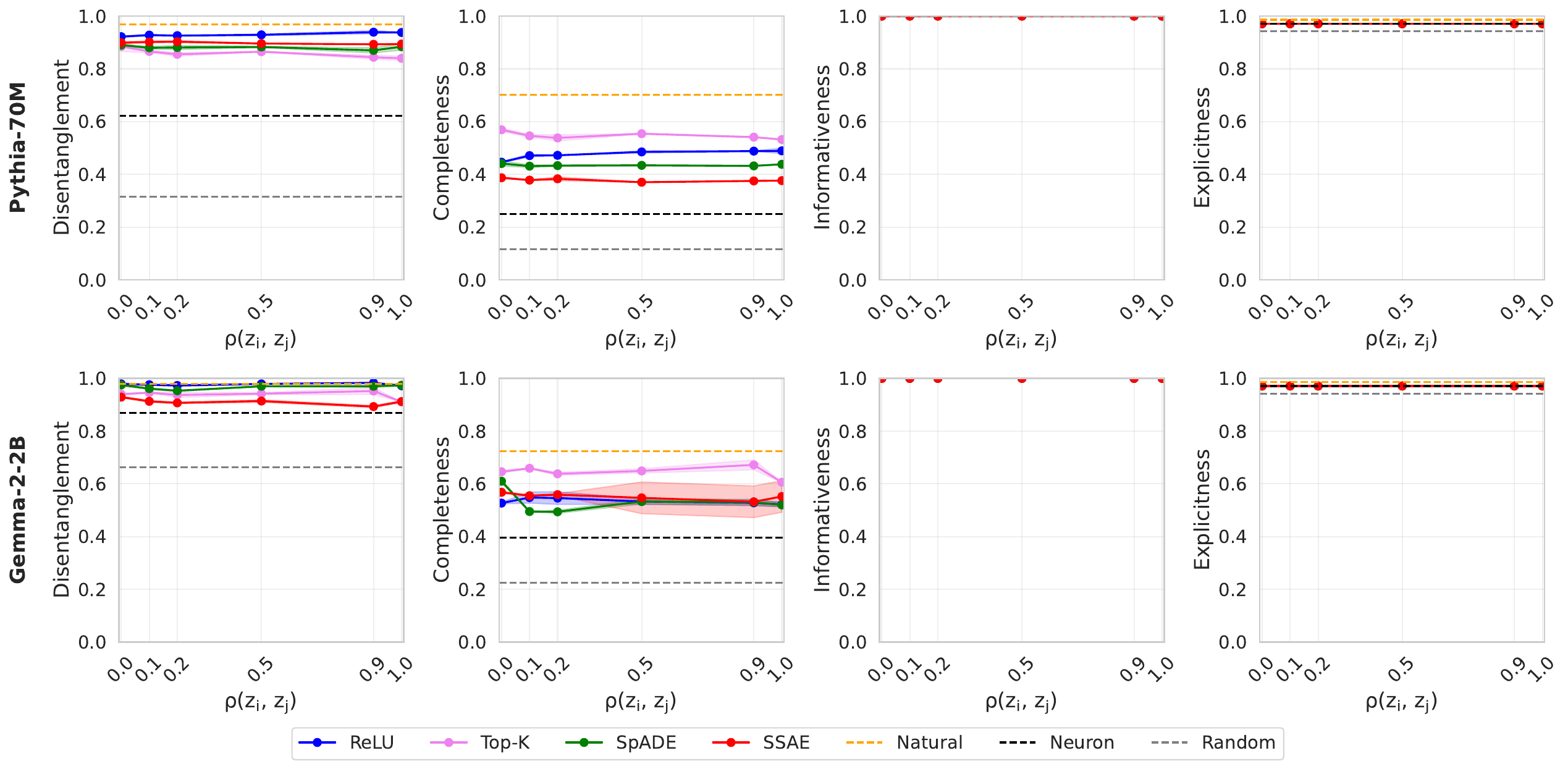}
    \caption{\textbf{DCI-ES scores under varying correlational conditions}. Shaded regions represent 1 std.\ dev.\ across 3 training seeds. Ideal performance looks like a flat line at 1 for all metrics. All methods achieve high disentanglement, informativeness, and explicitness, but relatively low completeness. This suggests that most features capture only one concept, but also that concepts are generally distributed across multiple features.}
    \label{fig:dci_es}
\end{figure*}

SSAEs perform best or close to best among (S)SAEs, as hypothesized.
Top-K SAEs also perform well, although they underperform for sentiment=positive. Our SAEs trained on synthetically generated data achieve comparable performance to SAEs trained on a much larger natural language corpus (the \emph{Natural} SAEs in Figure~\ref{fig:identifiability}); SSAEs outperform both for Pythia-70M, but not for Gemma-2-2B. Other methods achieve comparable or lower performance. Thus, to locate concepts in language models, \emph{one may not need to worry about curating concept-specific data if one's dataset is sufficiently large.} But also, the SAE architecture makes a significant difference.

When do correlations between concepts start to impede concept identification? The answer depends on the method: probes and SpADE \citep{costa2025flathierarchicalextractingsparse} maintain consistent MCCs up to correlations of 0.5 between concept pairs in the training data. Beyond this, concept representations degrade. For SSAEs, MCC remains consistent up to complete correlations of 1.0, as its theory predicts \citep{joshi_identifiable_2025}. Overall, these results suggest that if one uses an optimal architecture, one may not need to be concerned about spurious correlates of a target concept, unless that correlation is near-complete.

\subsection{Diagnosing Failures in Identifiability}\label{sec:exp_dcies}

As seen in \S\ref{sec:exp_identifiability}, MCCs are far from ideal. To diagnose the cause, we conduct a more fine-grained evaluation.

\paragraph{Metrics.} We use the DCI-ES framework of \citet{eastwood2023dcies}. Specifically, we use the \textbf{D}isentanglement, \textbf{C}ompleteness, \textbf{I}nformativeness, and \textbf{E}xplicitness metrics.
See \S\ref{sec:prelim} and App.~\ref{app:metrics} for intuitive and detailed definitions, respectively. In this setting, DCI-ES allows us to diagnose phenomena such as \emph{feature splitting} (low completeness), \emph{concept entanglement} (low disentanglement), the features not encoding the concept at all (low informativeness), or features encoding concepts in a difficult-to-recover way (low explicitness).

DCI-ES can indicate whether and to what extent (or equivalence class) identifiability is achieved. Identifiability up to \textit{invertible linear transformations} is achieved if $I = E = 1$;
up to permutation and element-wise reparametrization if $D = C = I = 1$; 
and \textit{up to sign and permutation} if $D = C = I = E = 1$. 
Importantly, steering is not guaranteed to work when $I = E = 1$, as for steering, we select the single most correlated dimension, which can be a linear mixture of multiple concepts.
$D = C = I = E = 1$ implies that all concepts are encoded in single features, which means we could predict the impact of steering on concept probabilities via linear extrapolation---even under multiple steering operations.

\paragraph{Hypothesis.} We hypothesize that all concepts will be recoverable from SAE feature vectors---i.e., that $I$ will be near 1. Because SAEs are trained to be sparse, we expect $E$ to be close to 1. We also hypothesize that the main failure mode will be feature splitting (that is, one concept being split across many features). If so, $C$ should be low.

\paragraph{Results.} We observe (Figure~\ref{fig:dci_es}) that $D$, $I$, and $E$ are high for all SAE architectures, but not for the original representation space of the models. This suggests that all concepts are near-perfectly recoverable (high $I$) with limited-capacity classifiers (high $E$), and that each SAE identifies the ground-truth concepts up to linear transformation.  However, $C$ is low, which suggests widespread feature splitting, and that the SAEs do not identify concepts up to sign and permutation. Most features are sensitive to one concept (high $D$), but concepts are often distributed across many features (low $C$).

To what degree does feature splitting occur? To quantify, we use $k$-sparse probes \citep{gurnee2023finding} and analyze how many features are necessary before probe performance saturates. This generally requires 10 features; see App.~\ref{app:dimensionality}.

High $D, I,$ and $E$ suggest that steering should only affect the probability of the target concept being steered (i.e., that features will generally be selective for one concept).
In the following section, we test these predictions by steering the top SAE feature for each concept.

\section{Steering as a Causal Independence Test}\label{sec:exp_steer_eval}

\begin{figure*}
    \centering
    \includegraphics[width=0.8\linewidth,clip]{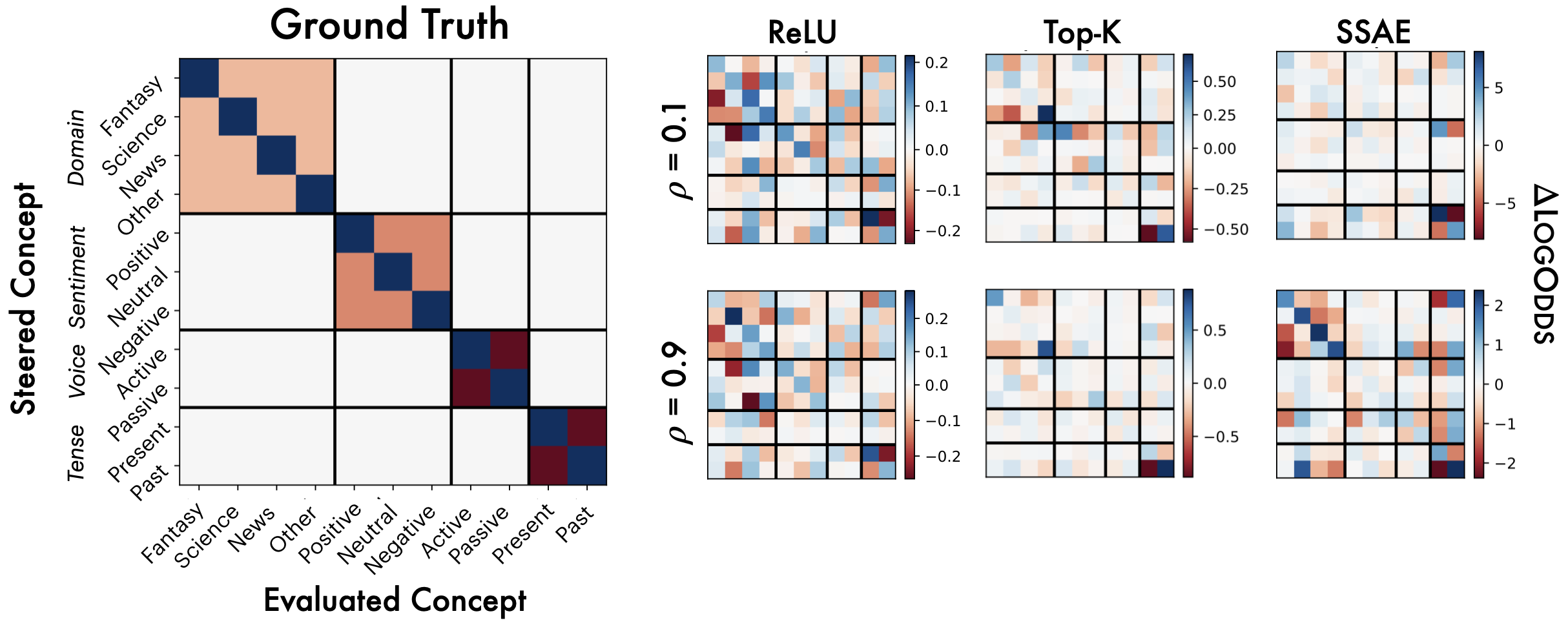}
    \caption{\textbf{The effect of steering a given concept (row) on the log-odds of another (column), as measured by a probe}. Results for Pythia-70M shown here; see App.~\ref{app:additional_steering_results_gemma2} for Gemma-2-2B. If concept representations are causally independent,
    we expect a heatmap that resembles the ground-truth: $\Delta$\textsc{LogOdds} should be high on the diagonal, negative for within-concept pairs, and close to 0.0 for across-concept pairs. All SAEs demonstrate the expected diagonals, but also significant across-concept effects. Increasing correlations in the training data, even up to 0.9, yield qualitatively similar entanglement patterns.}
    \label{fig:independence}
\end{figure*}

MCC and DCI-ES only provide correlational evidence.
However, none of these metrics provide \emph{causal} evidence that we can independently manipulate concepts using the learned features.
Thus, to measure causal efficacy, we employ steering as a test of independence of the mechanisms associated with each feature. This can be seen as testing the Independent Causal Mechanism principle~\citep{pearl_causal_2009,peters_elements_2018}, which holds that different causal mechanisms neither influence nor inform each other.

\subsection{Independence and Selectivity}\label{sec:exp_steer_heatmap}

We steer the top feature $\mathbf{\hat{f}}_i$ for concept $z_i$. To locate $\mathbf{\hat{f}}_i$, we select the feature whose activation correlates most with $z_i$, as when computing the MCC in \S\ref{sec:exp_identifiability}.\footnote{However, \citet{arad2025saesgoodsteering} has found that the features that \emph{detect} the input concept (the top-correlated features in our case) and the features that \emph{control} the concept in a model's outputs are nearly disjoint. Thus, we also test using gradient attributions \citep{simonyan2014deepinsideconvolutionalnetworks} to locate the feature that should be steered. See App.~\ref{app:additional_steering_results_gradient}.}

Steering the activations $\mathbf{h}^\ell$ with feature $\mathbf{f}_i$ is performed using a steering function \mbox{$\mathbf{\tilde{h}}^\ell(\mathbf{f}_i)\leftarrow \Phi(\mathbf{h}^\ell, \mathcal{F}, i, \alpha)$}, where $\Phi$ is defined as follows:
\begin{equation}
    \resizebox{\linewidth}{!}{
        $
        \Phi(\mathbf{h}^\ell,\mathcal{F},i, \alpha)=\mathcal{F}^{-1}\Big(\mathcal{F}(\mathbf{h^\ell})|\text{do}(\mathbf{f}_i=\alpha\cdot\text{max}(f_i))\Big) + \epsilon
        $
    }
\end{equation}
$\alpha$, the steering coefficient, controls the strength of the steering operation; $\mathcal{F}(\mathbf{h})$ corresponds to the featurized activations (equivalent to $\mathbf{f}$); and the do-operation denotes an intervention where feature $\mathbf{f}_i$ is set to $\alpha$ times its maximum activation $\max(f_i)$ on training dataset $\mathcal{D}$.
$\epsilon=\mathbf{h} - \mathcal{F}^{-1}(\mathcal{F}(\mathbf{h}))$ is the reconstruction error before interventions; adding it to the steered output ensures that any changes in model behavior are due to the steering operation, and not due to reconstruction errors \citep{marks2025sparse}. We set $\alpha$ to 5, but try multiple coefficients in \S\ref{sec:multisteer}.

\paragraph{Metrics.}
    For all concept pairs $(z_i,z_j)$, we steer the feature most associated with $z_i$ and measure $\Delta$\textsc{LogOdds} of concept $z_j$.
    We quantify $\Delta\textsc{LogOdds}(z_j)$ as the change in the logit of $z_j$ according to a multinomial concept
    probe trained on the \emph{final} layer of the model.\footnote{We use the final layer because it acts as a better proxy for the model's likely output behavior, as opposed to the model's inner representation of the input concepts. We use multinomial probes because they make the change in probabilities for within-concept pairs sum to 1. To validate that multinomial probe logits are good proxies for concept presence, we show heatmaps of probe accuracies in Figure~\ref{fig:probe_accuracy_heatmap_multi} (App.~\ref{app:probe_acc}).}
    
    We introduce \textbf{concept independence} $\mathcal{I}_S$ to quantify to what degree a concept is influenced only by its top feature and no others, and \textbf{feature selectivity} $\mathcal{S}_S$ to quantify to what degree a feature only influences its respective concept. To measure both, we construct a matrix $S \in \mathbb{R}^{Z\times Z}$, where rows correspond to steering $\hat{\mathbf{f}_i}$, the top feature for concept $i$, and columns correspond to $\Delta$\textsc{LogOdds}($z_j$). Then, for concept $z_i$, $\mathcal{I}_S$ is $S_{i,i}$ divided by the sum over column $i$. $\mathcal{S}_S$ is defined as $S_{i,i}$ divided by the sum over row $i$. More formally:
    \begin{align}
        \mathcal{I}_S &= \frac{|\log p(z_{\color{red}i} |\mathbf{\tilde{h}}^\ell(\mathbf{\hat{f}}_{\color{red}i})) - \log p(z_{\color{red}i})|}{\sum_{{\color{blue}j}\neq {\color{red}i}} (|\log p(z_{\color{red}i}|\mathbf{\tilde{h}}(\mathbf{\hat{f}}_{\color{blue}j})) - \log p(z_{\color{red}i})|)}, \\
        \mathcal{S}_S &= \frac{|\log p(z_{\color{red}i} |\mathbf{\tilde{h}}^\ell(\mathbf{\hat{f}}_{\color{red}i}))-\log p(z_{\color{red}i})|}{\sum_{{\color{blue}j}\neq {\color{red}i}} (|\log p(z_{\color{blue}j}|\mathbf{\tilde{h}}(\mathbf{\hat{f}}_{\color{red}i}))-\log p(z_{\color{blue}j})|)},
    \end{align}
    In words, $\mathcal{I}_S$ is maximized when steering $\mathbf{\hat{f}}_i$ affects $\log p(z_i)$ far more than steering any $\mathbf{\hat{f}}_j$, the top features for other concepts $z_j$. $\mathcal{S}_S$ is maximized when steering $\mathbf{\hat{f}}_i$ affects $\log p(z_i)$ far more than it affects any $\log p(z_j)$.
    In both equations, $j$ excludes all values of concept $z_i$. For example, if $i$ is domain=news, $j$ would skip all domains, including fantasy, news, etc.

    In practice, we use $\Delta$\textsc{LogOdds} rather than log-probability differences, as they are more likely to be additive at especially high and low probabilities.

\begin{table}[t]
    \resizebox{\linewidth}{!}{
    \begin{tabular}{lrrrrr}
    \toprule
    & & \multicolumn{2}{c}{Pythia-70M} & \multicolumn{2}{c}{Gemma-2-2B}\\\cmidrule(lr){3-4}\cmidrule(lr){5-6}
    SAE & $\rho$ & Independence & Selectivity & Independence & Selectivity \\
    \midrule
    \multirow{2}{*}{ReLU} & 0.1 & 0.32 (\textbf{0.53}) & 0.29 (\textbf{0.38}) & 0.24 (\textbf{0.49}) & 0.25 (\textbf{0.36}) \\
        & 0.9 & 0.30 (\textbf{0.40}) & 0.31 (\textbf{0.41}) & 0.23 (\textbf{0.62}) & 0.29 (\textbf{0.52}) \\
     \midrule
    \multirow{2}{*}{Top-K} & 0.1 & 0.60 (\textbf{0.86}) & 0.84 (\textbf{1.01}) & 0.43 (\textbf{0.65}) & 0.45 (\textbf{0.60}) \\
        & 0.9 & 0.49 (\textbf{0.55}) & 0.53 (\textbf{0.76}) & 0.76 (\textbf{1.25}) & 0.67 (\textbf{0.85}) \\
    \midrule
    \multirow{2}{*}{SSAE} & 0.1 & 0.42 (\textbf{0.65}) & 0.29 (\textbf{0.50}) & 0.50 (\textbf{1.67}) & 0.37 (\textbf{0.80}) \\
        & 0.9 & 0.37 (\textbf{0.71}) & 0.50 (\textbf{1.29}) & 0.72 (\textbf{1.21}) & 0.36 (\textbf{0.83}) \\
    \bottomrule
    \end{tabular}}
    \caption{\textbf{Steering independence and steering selectivity scores.} We present mean scores per feature/concept, and maxima across features/concepts in parentheses and bold. High independence means that a concept is only influenced by one feature; high selectivity means that a feature only influences one concept. Mean independence and selectivity are generally low, indicating widespread entanglement; however, maximal scores are high, indicating that at least one concept is selectively recovered by these architectures.}
    \label{tab:steer_select}
\end{table}

\paragraph{Hypothesis.} Because we observed high disentanglement in \S\ref{sec:exp_correlational}, we expect high feature selectivity and high concept independence. Because we observed low completeness, failures in steering should be because steering fails to affect its target concept, and not because a feature significantly affects unintended concepts.

\begin{figure*}
    \centering
    \includegraphics[width=0.9\linewidth]{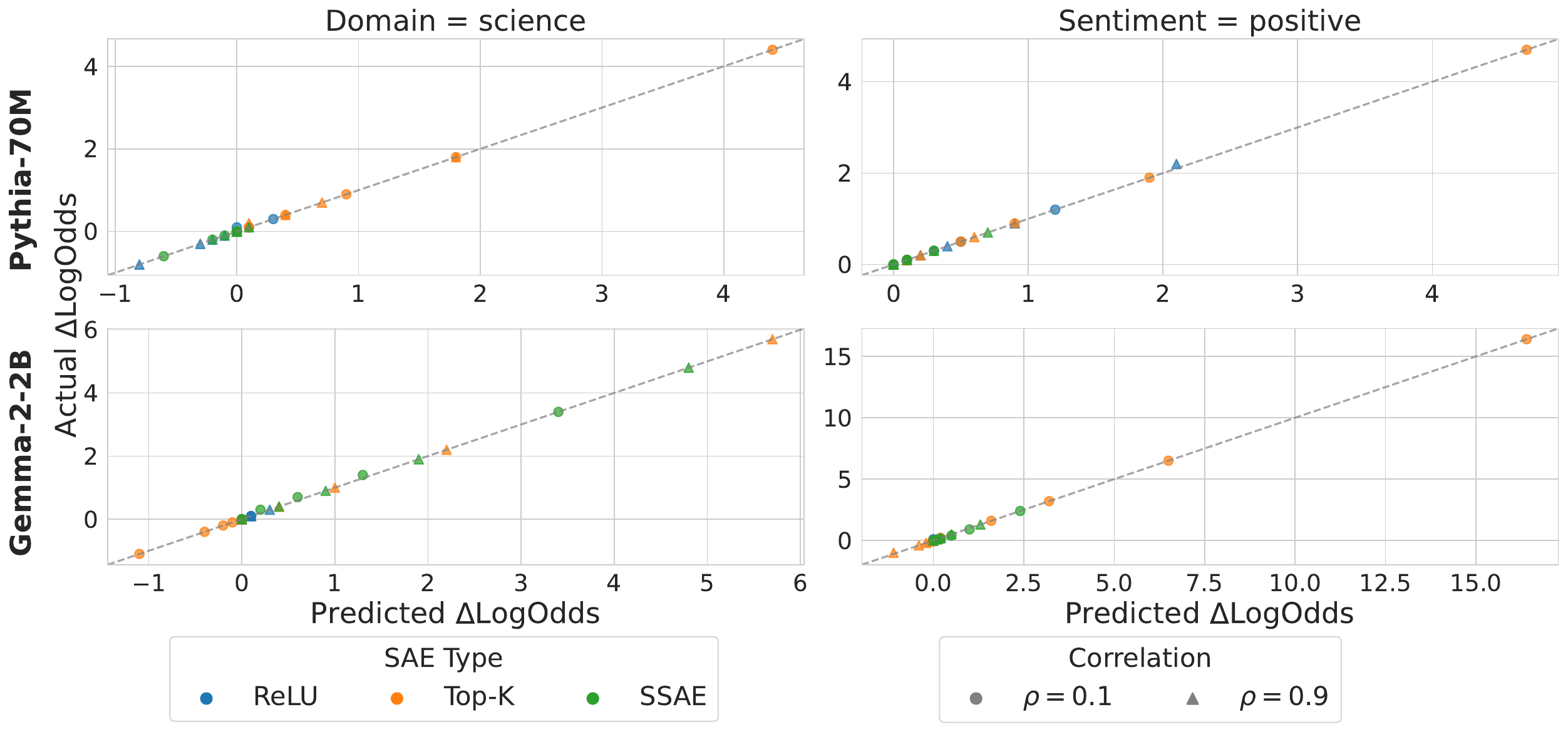}
    \caption{\textbf{Predicted $\Delta$\textsc{LogOdds}($z_i$) under disjointness assumptions vs.\ actual $\Delta$\textsc{LogOdds}($z_i$) when steering relevant feature $\mathbf{\hat{f}}_i$ and unrelated feature $\mathbf{\hat{f}}_j$.} Predicted $\Delta$\textsc{LogOdds} are obtained by adding the $\Delta$\textsc{LogOdds}($z_i$) when steering with either  $\mathbf{\hat{f}}_i$ or $\mathbf{\hat{f}}_j$ in separate forward passes. Actual $\Delta$\textsc{LogOdds} values are obtained by steering both in the same forward pass. $\mathbf{\hat{f}}_i$ and $\mathbf{\hat{f}}_j$ are typically disjoint, as indicated by the predicted change almost perfectly matching the actual change.}
    \label{fig:compositional_steering_pred_vs_actual}
\end{figure*}

\paragraph{Results.} We observe (Figure~\ref{fig:independence}) that for each SAE architecture, the expected diagonal trend is present, indicating that steering is increasing the log-odds of the target concept as expected. However, in even the best architectures, steering leads to measurable impacts on many unrelated concepts, indicating that selectivity is low. Moreover, concepts are often affected by many features, indicating widespread non-independence. Table~\ref{tab:steer_select} quantitatively summarizes these results. Best-case scores are high, but mean scores are lower, indicating that selective manipulability is achieved only for a  subset of concepts.

This underscores the importance of both multi-concept evaluations \emph{and} counterfactual interventions in evaluating concept representations: the correlational analyses did not suggest that interference would be likely in a steering setup, and yet we find evidence of widespread non-selectivity/non-independence. This may extend the findings of \citet{arad2025saesgoodsteering}, who claim that ``input features'' (those detecting a concept in the inputs) and ``output features'' (those controlling the model's use of the concept in its outputs) are disjoint. If this is true, then disentangled input features may be irrelevant to our ability to selectively steer.

\subsection{Disjointness}\label{sec:multisteer}
Steering with one concept and evaluating across many others can provide causal evidence as to how disentangled two concept mechanisms are. Now, inspired by \citet{zuheng2024automateddiscoverypairwiseinteractions}, we ask whether these concept representations are \textbf{disjoint}---i.e., whether they affect non-overlapping subspaces, and have no non-linear interaction terms. 
This is not equivalent to independence nor selectivity:\footnote{See Figure~\ref{fig:cd} (App.~\ref{app:metrics}) for an illustration and discussion of the difference between independence and disjointness.} even if two features have no interaction terms, each could still produce non-zero effects on multiple concepts.

Disjointness implies that we can predict the effect of pairs of steering operations on $z_i$ from individual steering operations, even if individual steering operations affect multiple concepts. 
Studying disjointness is important because its presence gives us predictive power over model behavior, even on unseen or out-of-distribution data.
See Figure~\ref{fig:cd} for illustrations and a direct contrast of independence and disjointness.
Formally, disjointness is achieved when:
\begin{multline}
    \log p(z_{\color{red}i} |\mathbf{\tilde{h}}^\ell(\mathbf{\hat{f}}_{\color{red}i},\mathbf{\hat{f}}_{\color{blue}j})) - \log p(z_{\color{red}i} | \mathbf{h}^\ell) =\\ \Big(\log p(z_{\color{red}i}|\mathbf{\tilde{h}}^\ell(\mathbf{\hat{f}}_{\color{red}i})) - \log p(z_{\color{red}i}|\mathbf{h}^\ell)\Big) +\\ \Big(\log p(z_{\color{red}i}|\mathbf{\tilde{h}}^\ell(\mathbf{\hat{f}}_{\color{blue}j})) - \log p(z_{\color{red}i}|\mathbf{h}^\ell)\Big).
\end{multline}
That is, the change in $\log p(z_i)$ when steering both $\mathbf{\hat{f}}_i$ and $\mathbf{\hat{f}}_j$ in one forward pass should be equivalent to the sum of the changes when steering $\mathbf{\hat{f}}_i$ and $\mathbf{\hat{f}}_j$ in separate forward passes. As in the previous experiment, we again use $\Delta$\textsc{LogOdds} rather than probabilities, as they are more likely to be additive at especially high and low probabilities.

We try steering coefficients $\alpha\in$ \{0.1, 0.5, 1.0, 2.0, 5.0\}. To compute predicted $\Delta$\textsc{LogOdds}, we steer $\mathbf{\hat{f}}_i$ and $\mathbf{\hat{f}}_j$ in separate forward passes and sum their effects on the \textsc{LogOdds} of concept $z_i$. To compute actual $\Delta$\textsc{LogOdds}, we steer both features in the same forward pass and measure the effect on $z_i$. If the prediction is equal to the actual change, we say that $\mathbf{\hat{f}}_i$ and $\mathbf{\hat{f}}_j$ are disjoint.

\paragraph{Hypothesis.} Under low correlations, we expect that features will be disjoint, such that the effect of steering the top features for $z_i$ and $z_j$ on $\Delta$\textsc{LogOdds}$(z_i)$ will be additive, regardless of the concepts' (non-)independence. Under higher correlations between concepts in the data, we expect less disjoint concept representations.

\begin{table}[t]
    \resizebox{\linewidth}{!}{
    \begin{tabular}{lrrrrr}
    \toprule
    & & \multicolumn{2}{c}{Domain=science} & \multicolumn{2}{c}{Sentiment=positive}\\\cmidrule(lr){3-4}\cmidrule(lr){5-6}
    SAE & $\rho$ & Pythia-70M & Gemma-2-2B & Pythia-70M & Gemma-2-2B \\
    \midrule
    \multirow{2}{*}{ReLU} & 0.1 & 1.00 & 0.99 & 1.00 & 1.00 \\
        & 0.9 & 1.00 & 1.00 & 1.00 & 1.00 \\
     \midrule
    \multirow{2}{*}{Top-K} & 0.1 & 1.00 & 0.99 & 1.00 & 0.98 \\
        & 0.9 & 1.00 & 1.00 & 1.00 & 0.99 \\
    \midrule
    \multirow{2}{*}{SSAE} & 0.1 & 1.00 & 1.00 & 1.00 & 1.00 \\
        & 0.9 & 1.00 & 1.00 & 1.00 & 1.00 \\
    \bottomrule
    \end{tabular}}
    \caption{\textbf{$R^2$ between predicted and actual $\Delta$\textsc{LogOdds}($z_i$) for each SAE.} Values are all near 1.00, indicating near-perfect disjointness for each SAE, even under high correlations between concepts.}
    \label{tab:disjointness}
\end{table}

\paragraph{Results.} We observe (Figure~\ref{fig:compositional_steering_pred_vs_actual}) that the effect of steering with two concepts simultaneously is almost exactly equivalent to summing the impact of steering with both concepts separately. To quantitatively verify, we compute the $R^2$ between predicted and actual $\Delta$\textsc{LogOdds}; these results (Table~\ref{tab:disjointness}) suggest almost no interaction.

This in combination with the non-independence results of \S\ref{sec:exp_steer_heatmap} suggests that feature interactions do not explain non-selective steering.
This suggests that the independence of two concept mechanisms cannot be established by demonstrating that their component sets or subspaces do not interact: one \emph{must} observe changes to output behaviors before and after intervention.

\section{Related Work}

\textbf{Featurization.} In interpretability, \emph{featurization} refers to techniques that allow one to map from less interpretable representations---typically activations---to more interpretable (and often sparser) representations (often called \emph{features}). This has produced supervised techniques such as sparse probing \citep{gurnee2023finding}, unsupervised techniques such as sparse autoencoders (SAEs; \citealp{olshausen-1997-sparse,bricken2023monosemanticity,huben2024sparse}), and non-parametric techniques such as steering vectors \citep{subramani-etal-2022-extracting} derived via difference-in-means \citep{marks2024the}.

How can one evaluate the quality of a feature? Recent work has proposed standardized evaluations based on known concepts \citep{mueller2025mib,huang-etal-2024-ravel,wu2025axbench}. These allow one to assess whether a concept discovery method can represent or enable counterfactual manipulation of a concept with high recall. Such studies suggest that SAEs are generally worse than simple supervised approaches like probes and difference-in-means. While \citet{wu2025axbench} and \citet{mueller2025mib} consider the quality of steering methods for one concept at a time, our benchmark additionally considers the relationship between the steered concept and others. We propose that evaluating disentanglement in steering requires multi-concept evaluations.

\textbf{Causal representation learning.}
Causal representation learning (CRL; \citealp{scholkopf_towards_2021}) assumes that high-dimensional observations are generated from low-dimensional latent factors, whose relationships to other latent factors are encoded in a causal graph. Then, CRL proposes latent variable models of such observations that are \textbf{identifiable}, meaning that the recovered features
are related to the true factors up to permutation and element-wise transformations. Because unsupervised learning is not identifiable without further assumptions \citep{hyvarinen_nonlinear_1999,darmois1951analyse,locatello_challenging_2019}, CRL methods rely on non-i.i.d.\ data or constraints on the decoding function \citep{moran_identifiable_2022,gresele_independent_2021,lachapelle_additive_2023,bradyinteraction,reizinger_jacobian-based_2023}.
For example, CRL has developed identifiable models using data from sparse interventions \cite{ahuja2023interventional,zhang_identifiability_2023,buchholz2023learning,von_kugelgen_nonparametric_2023}, contrastive pairs of samples \citep{ahuja2022weakly,locatello_weakly-supervised_2020, gresele_incomplete_2019,brehmer_weakly_2022}, data from multiple environments \citep{ahuja_multi-domain_2023,layne2025sparsity,khemakhem_variational_2020}, and temporal data with sparse or intervened mechanisms \citep{lachapelle_disentanglement_2021,lippe_biscuit_2023,lippe_citris_2022}. We go further, however, and test the causal implications of disentangled features on model outputs.
Similar to what we propose, \citet{joshi_identifiable_2025} propose a method that enables identifiable steering under multi-concept shifts; this method often performs well on disentanglement \textit{and} steering-based metrics.

To corroborate the theoretical claims of identifiability, access to the ground-truth factors is required, which generally limits the tasks that can be considered. Among the evaluation metrics, the MCC~\citep{hyvarinen_unsupervised_2016} has been used widely, despite known shortcomings \citep{hsu_disentanglement_2023}. Several other metrics have been proposed in both the disentanglement and the identifiable (causal) representation learning communities, such as the IRS score that measures interventional effects~\citep{suter2019robustly}, or the DCI~\citep{eastwood2018framework}, DCI-ES~\citep{eastwood2023dcies}, and InfoMEC~\citep{hsu_disentanglement_2023} scores that directly aim to address shortcomings of the MCC. See the concurrent work of \citet{joshi2026guardsguardianschallengesevaluating} for a detailed analysis.

\section{Discussion and Conclusions}
Our experiments reveal that current featurization methods, such as SAEs and sparse probes, show strong disentanglement when measured via correlational and representational evaluation metrics (\S\ref{sec:exp_identifiability}). Further analyses reveal that any failures in identifiability are usually due to the existence of many features for one concept, and not due to entanglement of concepts in individual features (\S\ref{sec:exp_dcies}). We also observe improvements in sparsity and disentanglement over the native neuron-based representation space of a model.

Even so, steering experiments reveal that entanglement in the output space can be widespread (\S\ref{sec:exp_steer_heatmap}, \ref{sec:multisteer}), even when correlational measures of disentanglement suggest otherwise. This extends findings from \citet{arad2025saesgoodsteering} to a multi-concept evaluation setting: disentangled input representations do not imply selective manipulability.

Despite the non-selectivity of features and non-independence of concepts during steering, most feature pairs demonstrate negligible interaction effects (\S\ref{sec:multisteer}) and operate over disjoint subspaces. This implies that when features achieve the \emph{form} of separation---that is, that they have no non-linear interactions in the activation spaces they correspond to---this does not necessarily imply that their \emph{functional roles} are non-interacting. This suggests that interpretability studies aiming to establish the independence of two mechanisms cannot settle for establishing that their subspaces or circuits do not interact; one must establish independence by observing the output behaviors of a system before and after interventions to those mechanisms. This could imply that circuit overlap is a poor proxy for mechanistic independence (or at least that it is not sufficient). It could also imply that featurizers that optimize for feature orthogonality may not actually be recovering independent mechanisms. Future work could focus on designing methods that directly optimize for counterfactual independence; for example, causal abstraction \citep{geiger_causal_2024} may provide a strong foundation for disentangling concept representations such that independence across concept representations is maximized.

\section*{Limitations}
Our data is generated by a PCFG. While it is natural language, it is still a far narrower distribution of text compared to the distributions on which sparse autoencoders are normally trained or evaluated.

Relatedly, our dataset is relatively simple in that it considers four concepts. At a high level, we see two complementary research needs in this space: the first focuses on understanding why and under what conditions steering might fail, and the second measures how often features are likely to fail in practice. Our work is aligned with the first need, whereas larger-scale benchmarks like AxBench \citep{wu2025axbench} and MIB \citep{mueller2025mib} are aligned with the second. Our choice to focus on a smaller range of concepts allows us to precisely evaluate when and why concepts can be identified and steered by allowing us to create perfect ground-truth labels for many factors for each input. Nonetheless, we acknowledge that our conclusions could be strengthened by extending these experiments to a larger number of concepts. Relatedly, we also only study categorical concepts. Extending this framework to continuous concepts could reveal interesting new trends.

\section*{Acknowledgments}
This work was initiated at the Fourth Bellairs Workshop on Causality, held at the McGill University Bellairs Research Institute (14--21 February, 2025). We are grateful to the organizers, Perouz Taslakian and Alexandre Drouin, for enabling this collaboration, and to the participants of the workshop for many thoughtful discussions during the ideation phase of this project. We thank Zhijing Jin, Vitória Barin Pacela, Victor Veitch, Atticus Geiger, Kartik Ahuja, Frederick Eberhardt, and Thomas Icard for extended discussions on earlier versions of these ideas. We also thank Sankaran Vaidyanathan for constructive feedback.

This work was supported by a grant from Coefficient Giving to Aaron Mueller. Patrik Reizinger acknowledges his membership in the European Laboratory for Learning and Intelligent Systems (ELLIS) PhD program and thanks the International Max Planck Research School for Intelligent Systems (IMPRS-IS) for its support.

\bibliography{references,references2}

\appendix
\section{Data Generation}\label{app:data}
We use probabilistic context-free grammars (PCFGs) to generate the training data for our SAEs. Non-terminals have attributes corresponding to the ground-truth concepts. In Figure~\ref{fig:cfg}, we show a subsample of the rules in the grammar. Note that this sample is simplified: most terminal-generating rules have over 10 non-terminals, and there are more sentence templates than displayed in the figure.

Concepts are uniformly distributed by default, with approximately $\frac{T}{V}$ examples per concept value, where $V$ is the number of values per concept, and 
$T$ is the dataset size. When there are cross-concept correlations, the correlated concepts are upsampled. For example, correlating domain=science with sentiment=positive means sentiment is sampled uniformly for non-science domains, but sentiment=positive is upsampled when domain=science. Examples are generated by uniformly sampling terminals conditioned on the sampled concept values.

\begin{figure*}[h]
\begin{lstlisting}[
    breaklines=true,
    backgroundcolor=\color{lightgray!10},
    basicstyle=\ttfamily\small,
    frame=single,
    xleftmargin=0pt,
    columns=flexible,
    literate={->}{$\rightarrow$\ }3
]
S[active, present] -> Subj V O | Subj V O PP | Adv, Subj V O
S[active, past] -> Subj V_past O | Subj V_past O PP | Adv, Subj V_past O
S[passive, present] -> O is V_pp by Subj | Adv, O is V_pp 
                              | O is being V_pp by Subj
S[passive, past] -> O was V_pp by Subj | Adv, O was V_pp 
                          | O had been V_pp by Subj
    
Subj[news, positive] -> the successful team  | the innovative company
Subj[news, neutral] -> the government | the company
Subj[fantasy, negative] -> the evil sorcerer | the treacherous assassin 
V[negative] -> criticizes | condemns | rejects
V[neutral] -> announces | reports | explains
V_past[positive] -> celebrated | praised | endorsed
  
PP[neutral] -> in recent days | across different sectors
PP[positive] -> with remarkable success | beyond expectations
PP[negative] -> without proper justification | to widespread criticism 
\end{lstlisting}
\caption{Excerpts from the context-free grammar we use to generate the SAE training and evaluation datasets.}
\label{fig:cfg}
\end{figure*}

In Table~\ref{tab:example_sentences}, we show examples from our generated training set. When we generate without correlations between concepts, there is an approximately uniform distribution of each concept, and correlations of approximately 0 across all concept pairs. If a concept-value pair is correlated, we pre-compute the example set such that we can achieve the closest match to the desired correlation. When training SAEs, we iterate for multiple epochs over the full dataset (when there are no cross-concept correlations) or the subsampled dataset (when there are cross-concept correlations).
\begin{table*}[h]
\centering
\begin{tabular}{llllp{8cm}}
    \toprule
    \multicolumn{4}{c}{\textbf{Concept Label}} & \\\cmidrule(lr){1-4}
    Voice & Tense & Domain & Sentiment & \textbf{Example Sentence} \\
    \midrule
    Active & Present & Science & Positive & The brilliant scientist celebrates the remarkable findings. \\
    Active & Present & Science & Neutral & The expert announces the parameters in recent days. \\
    Active & Present & Science & Negative & As of today, the discredited theory rejects the inconclusive evidence. \\
    Active & Past & Fantasy & Negative & Unsuccessfully, the malevolent dragon damaged the corrupted land. \\
    Passive & Past & News & Neutral & The event was explained in the recent report. \\
    Passive & Present & Other & Positive & The pleasant surprise is endorsed advantageously by the talented artist. \\
    Passive & Past & Other & Neutral & The question was answered when the family announced the event. \\
    \bottomrule
\end{tabular} 
\caption{Examples of sentences generated by our PCFG.}
\label{tab:example_sentences}
\end{table*}

\section{SAE Training Details}\label{app:methods}
\subsection{SAE Architectures}
Here, we define sparse autoencoders and describe the differences between the architectures we study.

\paragraph{Sparse autoencoders.} The conceptually simplest architecture we deploy is the ReLU sparse autoencoder \citep{huben2024sparse,bricken2023monosemanticity}, which learns a mapping from $\mathbf{x} = \mathbf{h}^\ell$ to a learned sparse feature vector $\mathbf{f}$, and then reconstructs the activations $\mathbf{\hat{x}}$ given $\mathbf{f}$. More formally:
\begin{equation}
    \mathbf{f} = \text{ReLU}(W_\text{enc}\mathbf{x} + \mathbf{b}_\text{enc})
\end{equation}
\begin{equation}
    \mathbf{\hat{x}} = W_\text{dec}(\mathbf{f} - \mathbf{b}_\text{enc}) + \mathbf{b}_\text{dec}
\end{equation}
ReLU SAEs minimize $\mathcal{L} = \text{MSE}(\mathbf{x}, \mathbf{\hat{x}}) + \lambda\lVert\mathbf{f} \rVert_1$.

Top-K SAEs \citep{gao2025scaling} are similar to ReLU SAEs, but they strictly retain the top $k$ activations per sample and zero out all others:
\begin{equation}
    \mathbf{f} = \text{top-}k(W_\text{enc}\mathbf{x} + \mathbf{b}_\text{enc})
\end{equation}

Sparsemax distance encoders (SpADE) can capture nonlinearly separable and heterogeneous features; we refer readers to \citet{hindupur2025projectingassumptionsdualitysparse} for details. In formal terms:
\begin{equation}
    \mathbf{f} = \text{Sparsemax}(-\lambda d(\mathbf{x}, W))
\end{equation}
where $d(\mathbf{x}, W)_i= \lVert\mathbf{x}-W_i\rVert_2^2 $. \citet{hindupur2025projectingassumptionsdualitysparse} show that this architecture can capture more irregular concept geometries, whereas ReLU SAEs assume linear separability, and Top-K SAEs assume angular separability.

\paragraph{Sparse shift autoencoders.}
Sparse shift autoencoders (SSAEs; \citealp{joshi_identifiable_2025}) are trained using paired observations $(\x, \tilde\x)$ assumed to be sampled from the following generative process:
\begin{equation}
    S \sim p(S), \quad (\vec{c}, \tilde{\vec{c}}) \sim p(\vec{c}, \tilde{\vec{c}} \mid S),
\end{equation}
\begin{equation}
    \vec{x} \coloneqq g(\vec{c}), \quad \tilde{\vec{x}} \coloneqq g(\tilde{\vec{c}}) \,,
\end{equation}
where $S \subseteq \{1, \dots, d_c\}$ denotes the subset of concepts that vary between $\x$ and $\tilde\x$, and $d_c$ represents the dimension of \emph{varying concepts}, the concepts that are intervened upon in the dataset.

Note that SSAEs take as input \emph{difference vectors} $\deltaz \coloneqq f(\tilde\x) - f(\x) = \tilde\z - \z$ that represent concept differences in activation space and model them as:
\begin{flalign}
     \hatdeltac_V \coloneqq {r}(\deltaz) & \coloneqq \mathbf{W}_e (\deltaz - \mathbf{b}_d) + \mathbf{b}_e\,;\\
    \hatdeltaz \coloneqq {q}(\hatdeltac_V) & \coloneqq \mathbf{W}_d \hatdeltac_V + \mathbf{b}_d\, 
\end{flalign}
where $r:\mathbb{R}^{d_z} \rightarrow \mathbb{R}^{d_c}$ is an affine encoder  $q:\mathbb{R}^{d_c} \rightarrow \mathbb{R}^{d_z}$ is an affine decoder.
In words, the representation $r(\deltaz)$ predicts $\deltac_V$, i.e., the concept shifts corresponding to $\deltaz$.

SSAEs are trained to solve the following constrained problem:
\begin{align}
    (\hat{r},\hat{q
    }) \in \arg \min_{r,q} \mathbb{E}_{\x,\tilde\x} \left[ ||\deltaz - q(r(\deltaz))||^2_2\right]\ \label{eqn:recon}\\
    \text{s.t.}\ \ \mathbb{E}_{\x, \tilde\x} || r(\deltaz) ||_0 \leq \beta \label{eqn:sparse_constraint}\,,
\end{align}
where Eq.~\ref{eqn:recon} is the standard auto-encoding loss that encourages good reconstruction and Eq.~\ref{eqn:sparse_constraint} is a regularizer that encourages the predicted concept shift vector $\hatdeltac_V \coloneqq \hat r(\deltaz)$ to be sparse. Since the $\ell_0$-norm is non-differentiable, in practice we replace it by an $\ell_1$-norm leading to the following relaxed sparsity constraint:
\begin{flalign}
    \mathbb{E}_{\x, \tilde\x} || r(\deltaz)||_1 \leq \beta\,. 
    \label{eqn:sparse_constraint_l1}
    \end{flalign}
We then approximately solve this constrained problem by finding a saddle point of its Lagrangian using the ExtraAdam algorithm \citep{gidel2020variationalinequalityperspectivegenerative} as implemented by \citet{gallegoPosada2022cooper}. 

\subsection{Hyperparameters}
\paragraph{Sparse autoencoders.} All Pythia-70M sparse autoencoders are trained using a batch size of 128 sequences for 10000 steps. We train on the output of the middle layer (layer 3). Optimization is performed using Adam with an initial learning rate of $1\times 10^{-3}$, 200 warmup steps, and $\beta_1=0.9,\beta_2=0.95$. Top-$k$ SAEs are trained with $k=128$. For Gemma-2-2B, we use the same hyperparameters for all SAEs except SpADE, which has higher memory requirements; for this architecture, we reduce the batch size to 64 while maintaining all other hyperparameters.\footnote{We experimented with doubling the number of training steps to compensate for the halved batch size for Gemma-2-2B SpADE SAEs. Final loss reductions were very small, so we chose to continue using 10000 iterations for uniformity.} We also train on the middle layer (layer 13). Our implementation is based on that of \citet{hindupur2025projectingassumptionsdualitysparse}.

\paragraph{Sparse shift autoencoders.} For SSAEs, we must train on pairwise differences in activations. For this, we iterate over the training set to get example $x_i$, and then uniformly sample another example $x_j$, ensuring that $i\neq j$. Otherwise, we use similar hyperparameters as when training SAEs. Note that SSAEs should be trained on the \emph{final} layer of a model, rather than the middle layer: this choice is motivated by the claim that concepts in the output space are most easily linearly identified in the final layer \citep{joshi_identifiable_2025}.\footnote{Using different layers for different SAE architectures introduces a confound. However, in pilot experiments, we found that other architectures tended to yield worse disentanglement and steering results when trained on the final layer. Thus, the current locations seem to be closer to optimal than training all SAEs on the same layer.}

\begin{table}[]
    \centering
    \resizebox{\linewidth}{!}{
    \begin{tabular}{lrrrr}
    \toprule
    SAE Arch. & $\rho(z_i,z_j)$ & NMSE & Var.\ Explained & \% Sparsity \\
    \midrule
    \multirow{6}{*}{ReLU} & 0.0 & 0.004 (0.000) & 99.6 (0.0) & 58.7 (0.2) \\
         & 0.1 & 0.006 (0.000) & 99.7 (0.0) & 54.6 (0.1) \\
         & 0.2 & 0.006 (0.008) & 99.7 (0.0) & 54.4 (0.2) \\
         & 0.5 & 0.007 (0.000) & 99.7 (0.0) & 54.4 (0.1) \\
         & 0.9 & 0.003 (0.000) & 99.7 (0.0) & 54.4 (0.1) \\
         & 1.0 & 0.003 (0.000) & 99.7 (0.1) & 54.4 (0.1) \\
    \midrule
    \multirow{6}{*}{Top-K} & 0.0 & 0.058 (0.000) & 94.2 (0.0) & 97.6 (0.0) \\
         & 0.1 & 0.056 (0.000) & 94.4 (0.0) & 97.6 (0.0) \\
         & 0.2 & 0.056 (0.000) & 94.4 (0.0) & 97.6 (0.0) \\
         & 0.5 & 0.060 (0.000) & 94.0 (0.0) & 97.6 (0.0) \\
         & 0.9 & 0.057 (0.000) & 94.4 (0.0) & 97.6 (0.0) \\
         & 1.0 & 0.064 (0.000) & 93.6 (0.0) & 97.6 (0.0) \\
    \midrule
    \multirow{6}{*}{SpADE} & 0.0 & 0.003 (0.000) & 99.7 (0.0) & 58.8 (0.3) \\
         & 0.1 & 0.003 (0.000) & 99.7 (0.0) & 58.8 (0.2) \\
         & 0.2 & 0.003 (0.000) & 99.7 (0.0) & 59.3 (0.1)  \\
         & 0.5 & 0.003 (0.000) & 99.7 (0.0) & 60.6 (0.0) \\
         & 0.9 & 0.004 (0.000) & 99.6 (0.0) & 64.9 (0.1) \\
         & 1.0 & 0.005 (0.000) & 99.5 (0.0) & 66.8 (0.1) \\
    \midrule
    Natural & - & 0.005 & 99.5 & 99.8 \\
    \midrule\midrule
    \multirow{6}{*}{SSAE} & 0.0 & 0.004 (0.001) & 99.6 (0.0) & 98.8 (0.0) \\
         & 0.1 & 0.004 (0.001) & 99.6 (0.0) & 99.1 (0.0) \\
         & 0.2 & 0.005 (0.001) & 99.6 (0.0) & 99.0 (0.0) \\
         & 0.5 & 0.005 (0.001) & 99.6 (0.0) & 99.1 (0.0) \\
         & 0.9 & 0.005 (0.002) & 99.6 (0.0) & 99.0 (0.0) \\
         & 1.0 & 0.004 (0.001) & 99.6 (0.0) & 99.2 (0.0) \\
    \bottomrule
    \end{tabular}}
    \caption{Variance explained, losses, and sparsities for SAEs trained on the middle layer of Pythia-70M (or last layer in the case of SSAEs). SSAE results are not comparable to those of other SAEs; unlike other architectures, they are trained and evaluated on \textit{pairwise differences} of activations.}
    \label{tab:train_loss_pythia}
\end{table}

\begin{table}[]
    \centering
    \resizebox{\linewidth}{!}{
    \begin{tabular}{lrrrr}
    \toprule
    SAE Arch. & $\rho(z_i,z_j)$ & NMSE &  Var.\ Explained & \% Sparsity \\
    \midrule
    \multirow{6}{*}{ReLU} & 0.0 & 0.014 (0.000) & 98.7 (0.0) & 49.6 (0.1) \\
         & 0.1 & 0.014 (0.000) & 98.6 (0.0) & 49.6 (0.1) \\
         & 0.2 & 0.014 (0.000) & 98.7 (0.0) & 49.6 (0.0) \\
         & 0.5 & 0.014 (0.000) & 98.7 (0.0) & 49.9 (0.0) \\
         & 0.9 & 0.011 (0.000) & 98.9 (0.0) & 50.0 (0.0) \\
         & 1.0 & 0.010 (0.000) & 99.0 (0.0) & 50.0 (0.0) \\
    \midrule
    \multirow{6}{*}{Top-K} & 0.0 & 0.218 (0.001) & 78.1 (0.0) & 99.4 (0.0) \\
         & 0.1 & 0.218 (0.000) & 78.1 (0.0) & 99.4 (0.0) \\
         & 0.2 & 0.216 (0.001) & 78.3 (0.0) & 99.4 (0.0) \\
         & 0.5 & 0.218 (0.000) & 78.2 (0.0) & 99.4 (0.0) \\
         & 0.9 & 0.236 (0.000) & 76.4 (0.0) & 99.4 (0.0) \\
         & 1.0 & 0.269 (0.000) & 73.1 (0.0) & 99.4 (0.0) \\
    \midrule
    \multirow{6}{*}{SpADE} & 0.0 & 0.094 (0.001) & 90.6 (0.0) & 96.9 (0.1) \\
         & 0.1 & 0.091 (0.000) & 90.5 (0.0) & 96.8 (0.0) \\
         & 0.2 & 0.091 (0.001) & 90.4 (0.0) & 96.9 (0.0) \\
         & 0.5 & 0.099 (0.001) & 89.5 (0.0) & 96.7 (0.0) \\
         & 0.9 & 0.149 (0.000) & 84.4 (0.1) & 96.9 (0.0) \\
         & 1.0 & 0.167 (0.001) & 84.5 (0.1) & 96.2 (0.0) \\
    \midrule
    Natural & - & 0.064 & 93.6 & 99.6 \\
    \midrule\midrule
    \multirow{6}{*}{SSAE} & 0.0 & 0.064 (0.001) & 98.8 (0.0) & 91.9 (0.1) \\
         & 0.1 & 0.068 (0.000) & 98.8 (0.0) & 91.5 (0.0) \\
         & 0.2 & 0.061 (0.000) & 98.8 (0.0) & 91.6 (0.1) \\
         & 0.5 & 0.072 (0.000) & 98.8 (0.0) & 91.4 (0.0) \\
         & 0.9 & 0.069 (0.000) & 98.9 (0.0) & 91.3 (0.0) \\
         & 1.0 & 0.074 (0.001) & 99.0 (0.0) & 91.4 (0.0) \\
    \bottomrule
    \end{tabular}}
    \caption{Variance explained, losses, and sparsities for SAEs trained on the middle layer of Gemma-2-2B (or last layer in the case of SSAEs). SSAE results are not comparable to those of other SAEs; unlike other architectures, they are trained and evaluated on \textit{pairwise differences} of activations.}
    \label{tab:train_loss_gemma}
\end{table}

We present NMSE, variance explained, and percent sparsity on the test set in Table~\ref{tab:train_loss_pythia} (for Pythia) and Table~\ref{tab:train_loss_gemma} (for Gemma).

\paragraph{Probes.} All probes are logistic regression probes. The probes used in correlational experiments are trained on the middle layer of Pythia-70M or Gemma-2-2B for a maximum of 1000 steps. We use the implementation of \texttt{scikit-learn} \citep{scikit-learn}.\footnote{Specifically, we use the Newton-Cholesky solver.} $k$-sparse probes are identical in architecture and hyperparameters, but we filter the set of neurons or features to reduce dimensionality before training the probes (and also train them on featurized representations rather than the original activation space); see App.~\ref{app:sparse_probes} for details. For the binary probes, we balance the training dataset of each probe by uniformly subsampling the more frequent class such that the number of examples for both classes is the same.

For the multinomial probes used for evaluating steering, the architecture and hyperparameters are the same, except that the probe outputs one logit \emph{per concept value} rather than a single logit. These probes are trained on the final layer of Pythia-70M or Gemma-2-2B, as their purpose is to estimate the probability of a concept appearing in the model's output. Note that we do not rebalance the data for multinomial probes; we only  train multinomial probes on data where there are no cross-concept correlations, so there is already an approximately uniform distribution of labels for each probe's training set. 

\section{Identifiability Definitions}\label{app:sec_ident}
    Identifiability definitions formulate the permissible transformations---termed an equivalence class---of the learned latent factors $\mathbf{f}$ by such that the resulting probability distributions parametrized by the neural network are equivalent. The smaller the equivalence class, the stronger assumptions are generally required. 
    
\begin{definition}[Strong Identifiability~\citep{khemakhem_ice-beem_2020}]\label{def:strong_ident}
Given a parameter class $\Theta,$ when the feature extractors $\mathcal{F}_{\theta_1},\mathcal{F}_{\theta_2}$ produce latent representations $\mathbf{f}_1=\mathcal{F}_{\theta_1}(\mathbf{x}), \mathbf{f}_2=\mathcal{F}_{\theta_2}(\mathbf{x})$ from observations $\mathbf{x}$ that are equivalent up to scaled permutations and offsets $c$ for all $\theta_1, \theta_2 \in \Theta$, i.e.,
 \begin{align}
    \theta_1\sim \theta_2 \iff \mathbf{f}=\mathcal{F}_{\theta_1}(\mathbf{x}) =  \mathbf{D}\mathbf{P}\mathcal{F}_{\theta_2}(\mathbf{x})+c,
 \end{align}
where $\mathbf{D}$ is a diagonal and $\mathbf{P}$ a permutation matrix. Then $\theta_1, \theta_2$ fulfill an \textit{equivalence} relationship. 
\end{definition}

\begin{definition}[Weak Identifiability~\citep{khemakhem_ice-beem_2020}]\label{def:weak_ident}
Given a parameter class $\Theta,$ when the feature extractors $\mathcal{F}_{\theta_1},\mathcal{F}_{\theta_2} $ produce latent representations $\mathbf{f}_1=\mathcal{F}_{\theta_1}(\mathbf{x}), \mathbf{f}_2=\mathcal{F}_{\theta_2}(\mathbf{x})$ from observations $\mathbf{x}$ that are equivalent up to matrix multiplications and offsets $c$ for all $\theta_1, \theta_2 \in \Theta$, i.e.,
 \begin{align}
    \theta_1\sim \theta_2 \iff \mathbf{f}=\mathcal{F}_{\theta_1}(\mathbf{x}) =  \mathrm{\mathbf{A}}\mathcal{F}_{\theta_2}(\mathbf{x})+c,
 \end{align}
where $\mathrm{rank}(\mathbf{A})\geq \min\left(\dim\mathbf{f};\dim\mathcal{X}\right)$. Then $\theta_1, \theta_2$ fulfill an \textit{equivalence} relationship. 
\end{definition}

\begin{definition}[Identifiability up to elementwise nonlinearities~\citep{hyvarinen_nonlinear_2017}]\label{def:nl_ident}
Given a parameter class $\Theta,$ when the feature extractors $\mathcal{F}_{\theta_1},\mathcal{F}_{\theta_2}$ produce latent representations $\mathbf{f}_1=\mathcal{F}_{\theta_1}(\mathbf{x}), \mathbf{f}_2=\mathcal{F}_{\theta_2}(\mathbf{x})$ from observations $\mathbf{x}$ that are equivalent up to elementwise nonlinearities, matrix multiplications and offsets $c$ for all $\theta_1, \theta_2 \in \Theta$, i.e.,
 \begin{align}
    \theta_1\sim \theta_2 \iff \mathbf{f}=\mathcal{F}_{\theta_1}(\mathbf{x}) =  \mathrm{\mathbf{A}}\mathbf{\sigma}\left[\mathcal{F}_{\theta_2}(\mathbf{x})\right]+c,
 \end{align}
where $\mathrm{rank}(\mathbf{A})\geq \min\left(\dim\mathbf{f};\dim\mathcal{X}\right)$ and $\mathbf{\sigma}$ denotes an elementwise nonlinear transformation. Then $\theta_1, \theta_2$ fulfill an \textit{equivalence} relationship. 
\end{definition}

\section{Metrics}\label{app:metrics}

\subsection{MCC}
    Given a set of ground-truth concepts $\{z_1,\ldots,z_n\}$ that generate an input example $\mathbf{x}$ where each concept $z_j\in Z$, then $\forall i\in [1,\ldots,n]$, we compute $\mathbf{\hat{f}}_j= \arg\max_i|\rho_\mathcal{D}(f_i, z_j)|$, where $f_i$ is the activation of feature $\mathbf{f}_i$ and $\rho$ is the Pearson correlation.
Intuitively, $\mathbf{\hat{f}}_j$ is the 
feature whose activation correlates most with the value of $z_j$ on some training dataset $\mathcal{D}$. Given test set $\mathcal{T}$ where concepts are uniformly distributed w.r.t.\ each other (i.e., no built-in correlations), we use $\rho_\mathcal{T}(\mathbf{\hat{f}}_j,z_j)$ as a measure of how well the featurizer linearly identifies concept $z_j$. After locating the best features $\{\mathbf{\hat{f}}_j\}_{j=1}^n$ for each concept, we compute the MCC as the mean of their correlations with their respective concepts on $\mathcal{T}$. In other words:
\begin{equation}
    \text{MCC} = \frac{1}{n}\sum_{j=1}^n\rho_\mathcal{T}(\hat{f}_j, z_j).
\end{equation}

The MCC is measured using one-dimensional features, but multinomial concepts may not be one-dimensional in $\mathbf{f}$ or $\mathbf{h}^\ell$ \citep{engels2025not}. Thus, to create a fairer evaluation, we compute the MCC over binarized concepts. That is, given a variable $z_i\in Z$ with $V_i$ possible values, we create a new binary variable $v_{i,x}\in\mathbb{B}$ for each value $x$ corresponding to whether $z_i=v_{i,x}$. When computing the MCC, we first average the correlation coefficients for all $v_{i,x}\in V_i$ before taking the macroaverage across concepts.

\subsection{DCI-ES}
Here, we provide further detail on the DCI-ES metrics \citep{eastwood2023dcies}, and give methodological details as to how we compute them. Our implementation is based directly on that of \citet{eastwood2023dcies}.

DCI-ES stands for \textbf{d}isentanglement, \textbf{c}ompleteness, \textbf{i}nformativeness, \textbf{e}xplicitness, and \textbf{s}ize. We focus on the first four metrics, as these are the most relevant to establishing identifiability. Disentanglement and completeness require us to first compute importance matrix $M\in\mathbb{R}^{|\mathbf{f}|\times |Z|}$. For example, if we train a multinomial probe to predict concept $z_j$ from feature $\mathbf{f}_i$, we can compute the importance of each dimension of $\mathbf{f}$ post hoc. Each concept $z_j$ defines a column of $M$. Note that $\forall i,j:M_{ij}\geq 0,$ and $\sum_{i=1}^{|\mathbf{f}|}M_{ij}=1$.

\textbf{Disentanglement} measures the average number of concepts $z_j$ that are captured by any single feature $\mathbf{f}_i$. To compute it, we first compute the entropy $H_Z(P_{i.})$ of the distribution $P_{i.}$ defined over row $i$ of $M$: $P_{ij} = \frac{M_{ij}}{\sum_{k}M_{ik}}$. Disentanglement is then defined as $D_i=1-H_K(P_{i.})$. This score is maximized when feature $\mathbf{f}_i$ is only responsible for predicting a single concept $z_j$; it is minimized when feature $\mathbf{f}_i$ is equally important for predicting all concepts.

\textbf{Completeness} measures the average number of features $\mathbf{f}_i$ that are useful in predicting a single concept $z_j$. This score is defined analogously to disentanglement, but over columns $j$ in $M$: we take $C_j = 1-H_\mathbf{f}(P_{.j})$. Completeness is maximized when only one feature $\mathbf{f}_i$ is helpful in predicting $z_j$, and it is minimized when all features are equally important in predicting the concept.

\textbf{Informativeness} is inversely proportional to the prediction error of a probe trained on the feature vector. In the implementation of \citet{eastwood2023dcies}, it is simply defined as the accuracy of a probe in predicting concept $z_j$ when trained on the feature vector $\mathbf{f}$. This captures whether a ground-truth concept is recoverable from the feature vector.

\textbf{Explicitness} is conceptually related to informativeness. $E$ captures the trade-off between the probe's capacity and the probe loss; this is measured as one minus the normalized area under the loss-capacity curve (AULCC); we refer readers to \citet{eastwood2023dcies} for details. This score is maximized when the lowest-capacity probe achieves the best loss, and thus that no excess capacity was required to fully recover a given concept.

\subsection{Further Details on Independence and Disjointness}
To illustrate the conceptual distinction between independence and disjointness, we present diagrams in Figure~\ref{fig:cd}. Intuitively, disjointness implies that the influence of one feature on the output does not interact with another, and thus that the effect of steering both features can be predicted from the result of steering either in isolation. Independence instead implies that steering with one concept would not affect how the model uses other concepts. Refer to \S\ref{sec:multisteer} for details.

\begin{figure}
    \centering
    \begin{tikzcd}[column sep=huge, row sep=huge]
        p(z_i|\mathbf{h}^\ell) \arrow[r, "{\Phi(\mathbf{h}^\ell,\mathcal{F},i,\alpha)}"] \arrow[d, "{\Phi(\mathbf{h}^\ell,\mathcal{F},j,\beta)}"'] 
         & p(z_i|\mathbf{\tilde{h}}^\ell(\mathbf{\hat{f}}_i)) \arrow[d, "{\Phi(\mathbf{h}^\ell,\mathcal{F},j,\beta)}"] \\
        p(z_i|\mathbf{\tilde{h}}^\ell(\mathbf{\hat{f}}_j)) \arrow[r, "{\Phi(\mathbf{h}^\ell,\mathcal{F},i,\alpha)}"']
         & p(z_i|\mathbf{\tilde{h}}^\ell( \mathbf{\hat{f}}_i, \mathbf{\hat{f}}_j))
    \end{tikzcd}
    
    \vspace{1cm}
    
    \begin{tikzcd}[column sep=large, row sep=large]
        p(z_j|\mathbf{h}^\ell) \arrow[r, red, "\text{\textcolor{black}{\ \ ${\Phi(\mathbf{h}^\ell,\mathcal{F},i,\alpha)}$}\ \ }", "\text{no change}"']
          & p(z_j|\mathbf{\tilde{h}}^\ell(\mathbf{\hat{f}}_i))
    \end{tikzcd}
    \caption{\textbf{The difference between disjointness and independence:} \textbf{(Top)} Two concepts $z_i$ and $z_j$ with feature representations $\mathbf{\hat{f}}_i$ and $\mathbf{\hat{f}}_j$, respectively, are disjoint if the top diagram commutes. \textbf{(Bottom)} If they are independent then there is no commutative relationship, as steering with $\mathbf{\hat{f}}_i$ should not affect $p(z_j)$.
    Intuitively, disjointness implies that two features have no interaction terms, and thus that the effect of steering of both can be predicted from the result of steering either in isolation.
    Feature selectivity and concept independence instead imply that steering with one concept's top feature does not affect how the model uses other concepts. Refer to \S\ref{sec:multisteer} for formulae and empirical details.}
    \label{fig:cd}
\end{figure}

\section{Is One Dimension Sufficient?}\label{app:dimensionality}
In SAE-based interpretability studies, it is common to steer with a single feature, regardless of how many features receive high attributions for a given task. This corresponds to the following assumption:

\textbf{Assumption:} \emph{Given binary concept $z_i$ and feature vector $\mathbf{f}$, one dimension $\mathbf{f}_i$ of $\mathbf{f}$ is sufficient to represent and control $z_i$ in $\mathcal{M}$.}

\begin{figure*}
    \centering
    \includegraphics[width=0.8\linewidth]{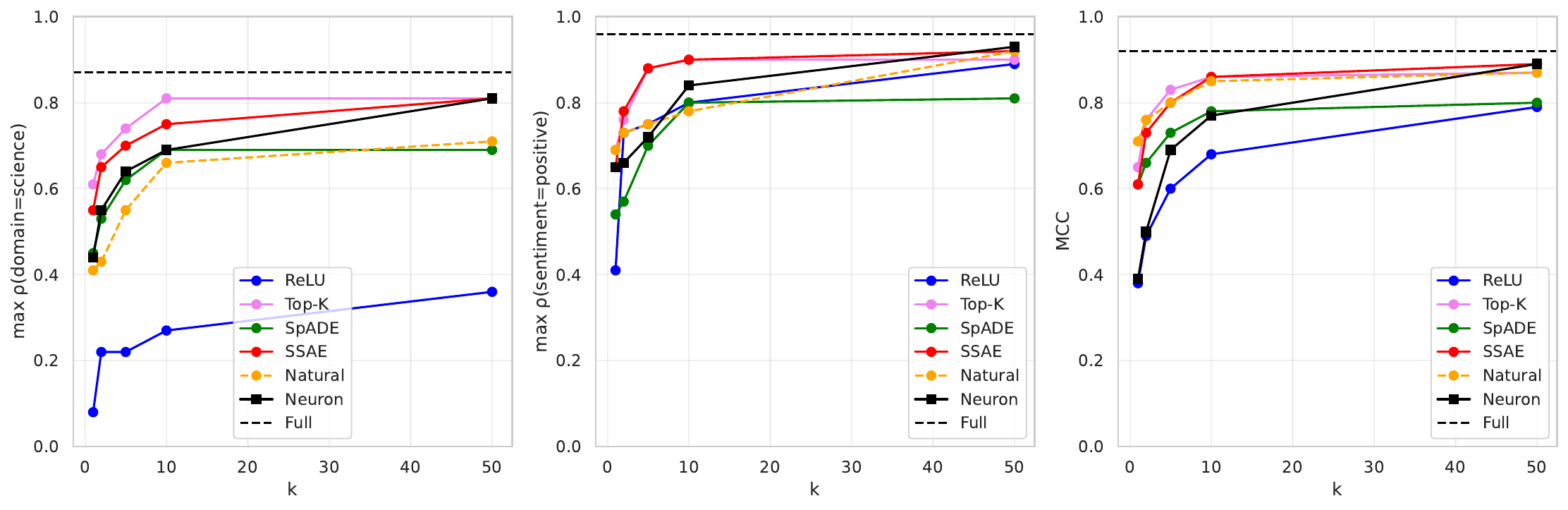}
    \caption{\textbf{Correlation coefficients between probe logits and concept labels for domain=science (left), sentiment=positive (middle), and MCC (right)}. Results for Gemma-2-2B shown here; results for Pythia-70M are in App.~\ref{app:disentanglement}. We vary the number of dimensions $k$ that the probe is allowed to have non-zero weights from. $k$-sparse probes trained on SAEs begin to converge around 10 dimensions for Top-K, SpADE, SSAE, and Natural, and recover most of the performance of a non-sparse probe that is allowed to use the entire residual vector (Full). $k$-sparse probes trained on the residual stream (Neuron) require more dimensions to converge, as expected.}
    \label{fig:sparsity}
\end{figure*}

To evaluate the extent to which this assumption holds in practice, we train $k$-sparse probes (as operationalized in \citet{gurnee2023finding}) on featurized representations $\mathbf{f}$.
$k$-sparse probes are linear probes that may have non-zero weights from up to $k$ dimensions of the representations they are trained on. \citet{lachapelle_synergies_2023}  established a connection between disentanglement and sparse prediction: they prove that disentanglement leads to optimal loss using sparse predictors.
Further, as features become more entangled, we need to reduce sparsity regularization to maintain accuracy; this theoretical finding further motivates the following experiment. 

\paragraph{Hypothesis.} More dimensions yield monotonically increasing expressive power. Thus, performance should be non-decreasing as $k$ increases. We care primarily about when increasing $k$ begins to yield diminishing improvements in the MCC. Representations obtained with strong sparsity constraints, like SAEs, should reach this saturation point at smaller $k$ than representations with no such constraints, such as residual vectors.

\paragraph{Results.}  The MCCs of $k$-sparse probes trained on feature vectors $\mathbf{f}$ are presented in Figure~\ref{fig:sparsity}. Top-K SAEs and SSAEs achieve the best trade-off between MCC and sparsity at all $k$; they also approach the MCC of training a normal probe on the full activation vector at the residual stream. ReLU SAEs do not begin saturating even at 10--50 features, whereas all other SAEs do. SSAEs and Top-K SAEs achieve better concept recovery at the same $k$ as the residual neuron baseline, whereas ReLU SAEs do not.

These results suggest that SAEs do confer sparsity benefits compared to the original activation space of $\mathcal{M}$, but also that one-dimensionality assumptions may often yield suboptimal results---even when the concepts are relatively simple. That said, the relative ordering of MCCs across architectures does not generally change with $k$, so comparisons across architectures should generally yield conclusions that are stable across choices of $k$ (at least up to $k=50$).

\section{Probe Accuracies}\label{app:probe_acc}
\begin{figure*}
    \centering
    \includegraphics[width=0.8\linewidth]{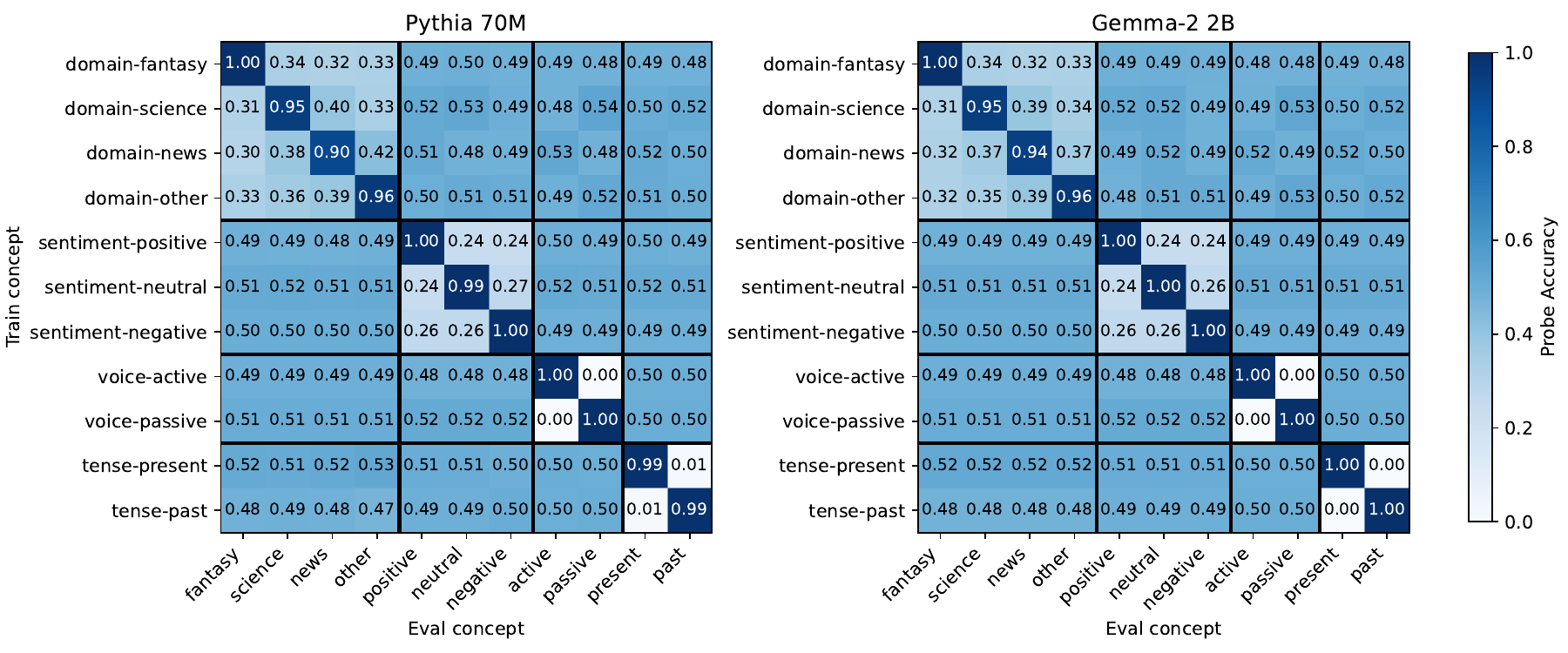}
    \caption{Accuracy of binary probes (rows) on all concept value classification tasks (columns). We expect high values on the diagonals, below random chance for within-concept value pairs, and random chance for across-concept value pairs.}
    \label{fig:probe_accuracy_heatmap_binary}
\end{figure*}

\begin{figure*}
    \centering
    \includegraphics[width=0.8\linewidth]{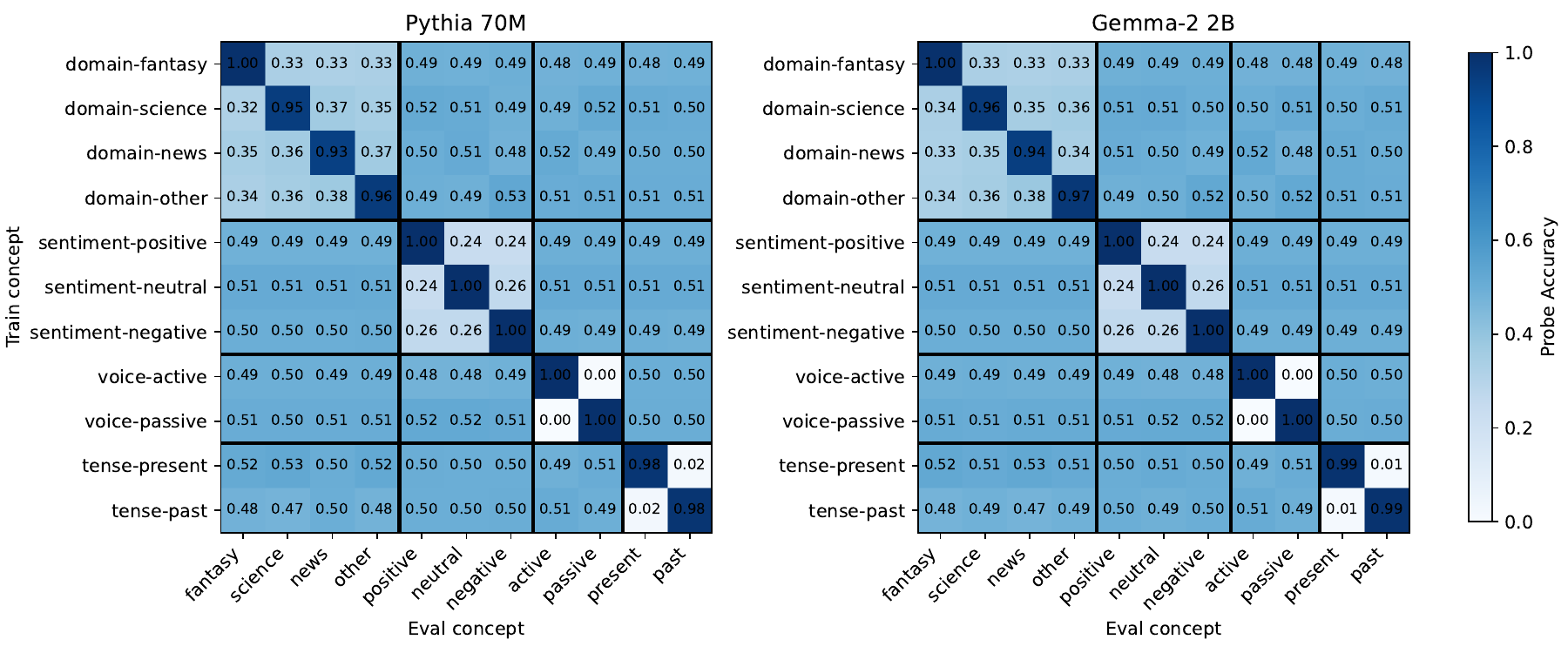}
    \caption{Accuracy of multinomial probes on all concept value classification tasks (columns). We expect high values on the diagonals, below random chance for within-concept value pairs, and random chance for across-concept value pairs.}
    \label{fig:probe_accuracy_heatmap_multi}
\end{figure*}

Here, we present the accuracies of each probe we use in our disentanglement experiments and evaluations. We present these as heatmaps to verify whether each probe learns an independent representation of its target concept; if it does, we expect high scores along the diagonal, lower-than-random scores for within-concept pairs,\footnote{ We expect lower-than-random scores for within-concept pairs because a classifier trained on an alternative value of a concept should be strictly worse than a random probe, as the target label will be \emph{negatively} correlated with the target concept.} and random-chance scores for across-concept pairs.

Binary linear probes trained on the middle layers of Pythia-70M and Gemma-2-2B (Figure~\ref{fig:probe_accuracy_heatmap_binary}) achieve near-perfect accuracies on their respective concepts, and achieve the expected random accuracies on all other concepts. This empirically supports Assumption 1, and supports the idea that the MCC ceiling should be high (\S\ref{sec:exp_identifiability}).

In \S\ref{sec:exp_steer_heatmap} and \S\ref{sec:multisteer}, we instead use multinomial linear probes trained on the final layers of Pythia-70M and Gemma-2-2B. We find (Figure~\ref{fig:probe_accuracy_heatmap_multi}) that these probes also achieve the expected high accuracies on the target concepts, below-random-chance accuracies on within-concept pairs, and random-chance accuracies on across-concept pairs. This validates that the non-independence we observe in our steering experiments are not due to the probes, but rather are more likely due to the featurization methods that we use to steer.

\section{Sparse Probing}\label{app:sparse_probes}
Here, we replicate the setup of \citet{gurnee2023finding} in our cross-concept correlation setting. We aim to assess which $k$-sparse probing methods are more robust to cross-concept correlations at multiple $k$. We focus on the two most performant methods from \citet{gurnee2023finding}: max mean difference (MD), and logistic regression (LR). MD works by computing the average difference in activations between positive and negative samples, and taking the $k$ neurons whose mean activation difference is greatest. LR works by first training a logistic regression probe with $L_1$ regularization on the full activation vector, and then taking the top $k$ according to the weights of the probe.

\begin{figure*}[t]
    \centering
    \includegraphics[width=0.9\linewidth]{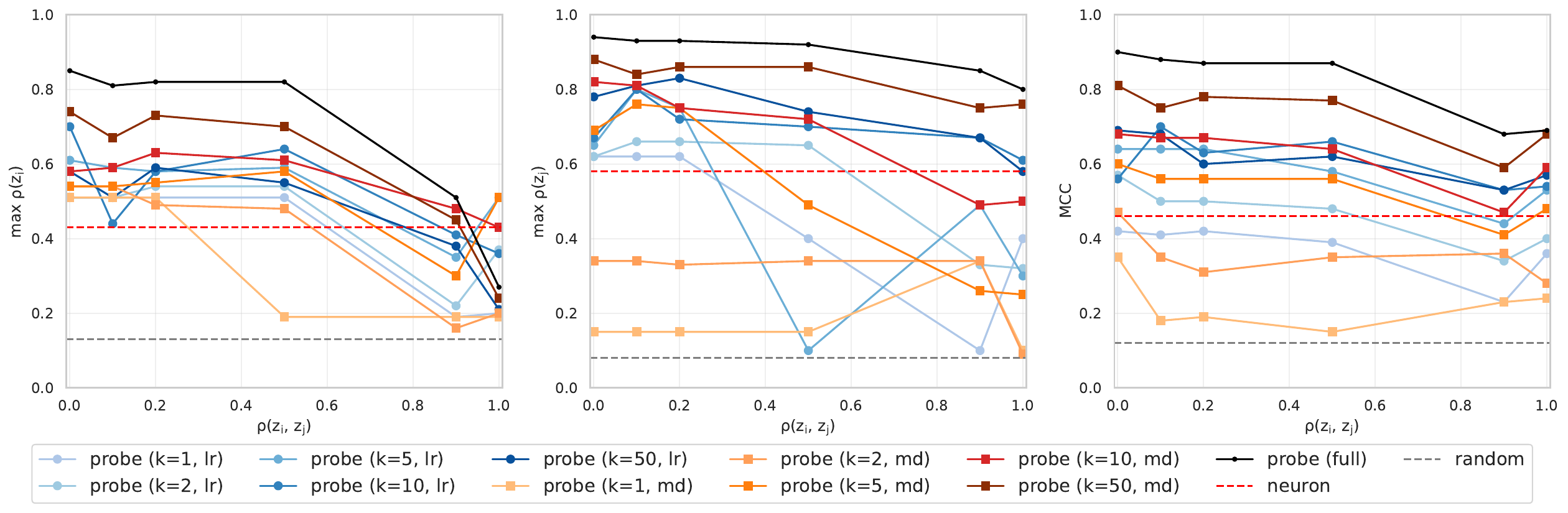}
    \caption{\textbf{MCC for the two most performant sparse probing methods from \citet{gurnee2023finding} at various $k$}. Results for Pythia-70M shown here. The LR method achieves higher MCC at lower $k$, but MD overtakes LR at higher $k$.}
    \label{fig:sparse_probing}
\end{figure*}

We observe (Figure~\ref{fig:sparse_probing}) that the logistic regression (LR) method of selecting neurons is more effective at lower $k$. Between $k=5$ and $k=10$, MD generally overtakes LR in performance. As we are more concerned with low-dimensional concept recovery, we focus on LR in the feature dimensionality experiment (\S\ref{app:dimensionality}).

\section{Further Disentanglement Results}\label{app:disentanglement}
\begin{figure*}
    \centering
    \includegraphics[width=0.9\linewidth]{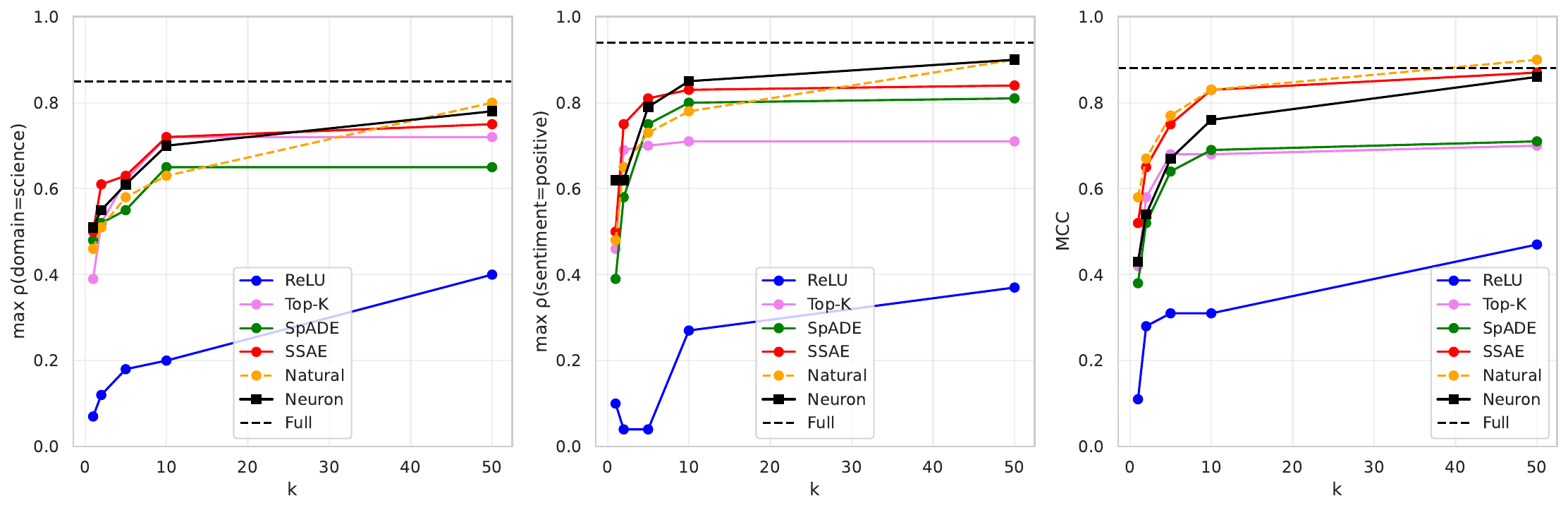}
    \caption{\textbf{Correlation coefficients between probe logits and concept labels for domain=science (left), sentiment=positive (middle), and MCC (right)}. Results for Pythia-70M. We vary the number of dimensions $k$ that the probe is allowed to have non-zero weights from. As with Gemma-2-2B, correlation coefficients tend to converge at around 10 dimensions. However, the neuron baseline is far more performant; perhaps this is because $k=10$ represents a far greater proportion of the dimensions of $\mathbf{h}^\ell$ for Pythia than Gemma. Other trends are largely consistent with Figure~\ref{fig:sparsity}.}
    \label{fig:correlation_pythia}
\end{figure*}

Here, we present correlation coefficients and MCCs for $k$-sparse probes trained with varying $k$ on SAEs for Pythia-70M (Figure~\ref{fig:correlation_pythia}). As with Gemma-2-2B, correlation coefficients tend to converge at around 10 dimensions; this suggests that the one-dimensionality assumption may not often hold in practice, even for much smaller models. Note also that the neuron baseline is far more performant for Pythia than Gemma; perhaps this is because $k=10$ represents a far greater proportion of the dimensions of $\mathbf{h}^\ell$ for Pythia than Gemma. Other trends are largely consistent with Figure~\ref{fig:sparsity}.

\section{Further Steering Results}
\label{app:additional_steering_results}

\subsection{Results for Gemma 2}\label{app:additional_steering_results_gemma2}
\begin{figure*}[t]
    \centering
    \includegraphics[width=0.8\linewidth]{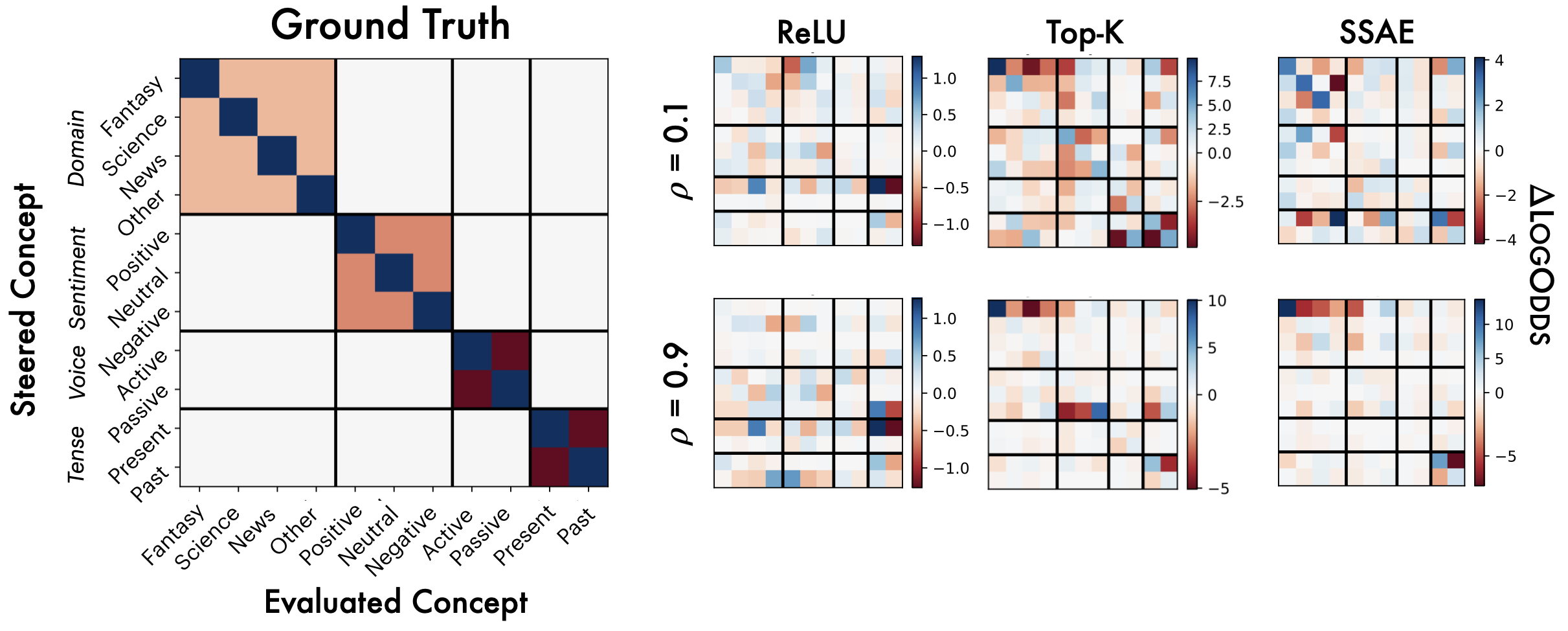}
    \caption{\textbf{The effect of steering a given concept (row) on the logit of another concept (column)}. Results for Gemma-2-2B. If concept representations are causally independent,
    we expect a heatmap that resembles the ground-truth: $\Delta$\textsc{LogOdds} should be high on the diagonal, negative for within-concept pairs, and close to 0.0 for across-concept pairs. All SAEs demonstrate the expected diagonals, but also significant across-concept effects, indicating non-independence. Results are consistent across low and high correlations between concepts in the SAEs' training data.}
    \label{fig:independence_gemma2}
\end{figure*}

Here, we present steering heatmaps for Gemma-2-2B (Figure~\ref{fig:independence_gemma2}). Features appear less independent than for Pythia-70M, as indicated by more significant across-concept $\Delta\text{LogOdds}$ for many concept pairs. That said, the expected diagonal trend is still present. This is further evidence that SAE features do not often correspond to causally independent concept representations.

\begin{figure*}[t]
    \centering
    \includegraphics[width=0.8\linewidth]{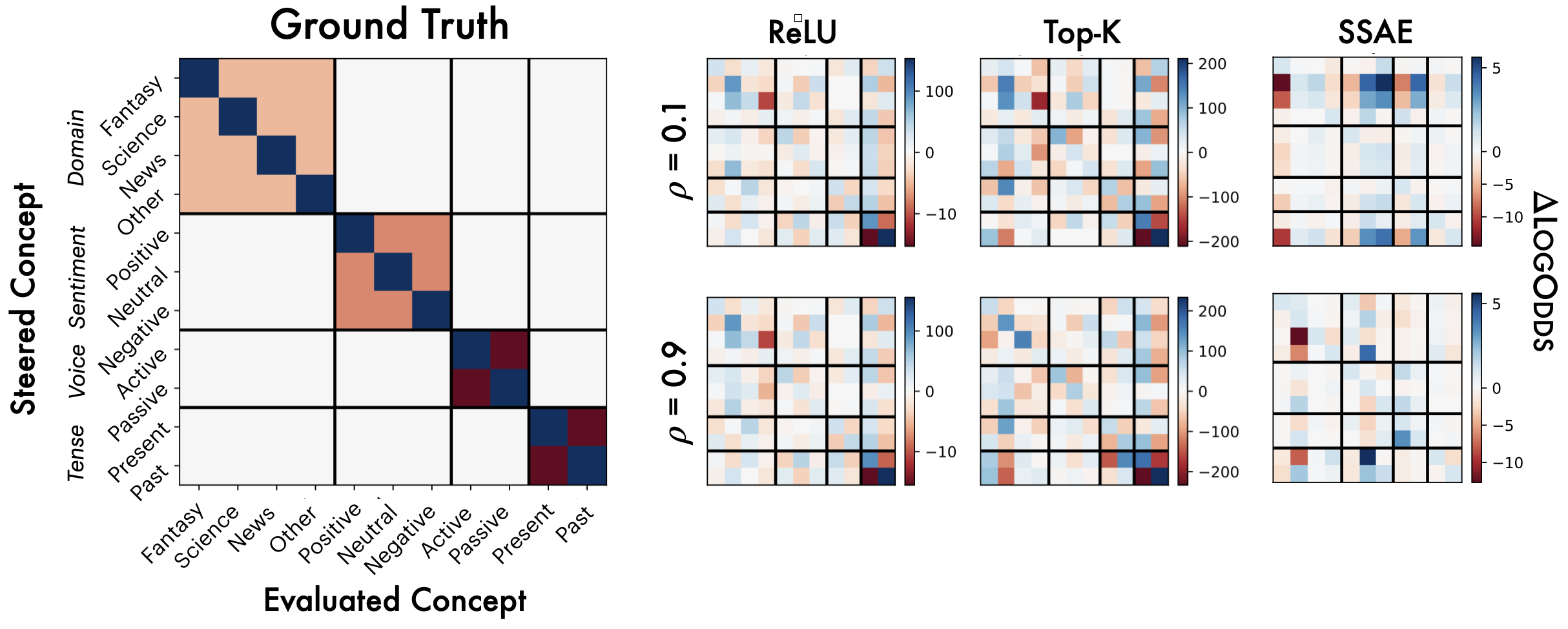}
    \caption{\textbf{The effect of steering a given concept (row) on the logit of another concept (column)}. Results for Pythia-70M when locating features using gradient attributions rather than activation correlations. If concept representations are causally independent,
    we expect a heatmap that resembles the ground-truth: $\Delta$\textsc{LogOdds} should be high on the diagonal, negative for within-concept pairs, and close to 0.0 for across-concept pairs. All SAEs demonstrate the expected diagonals, but also significant across-concept effects, indicating non-independence. Results are consistent across low and high correlations between concepts in the SAEs' training data.}
    \label{fig:independence_pythia_gradient}
\end{figure*}

\begin{figure*}[t]
    \centering
    \includegraphics[width=0.8\linewidth]{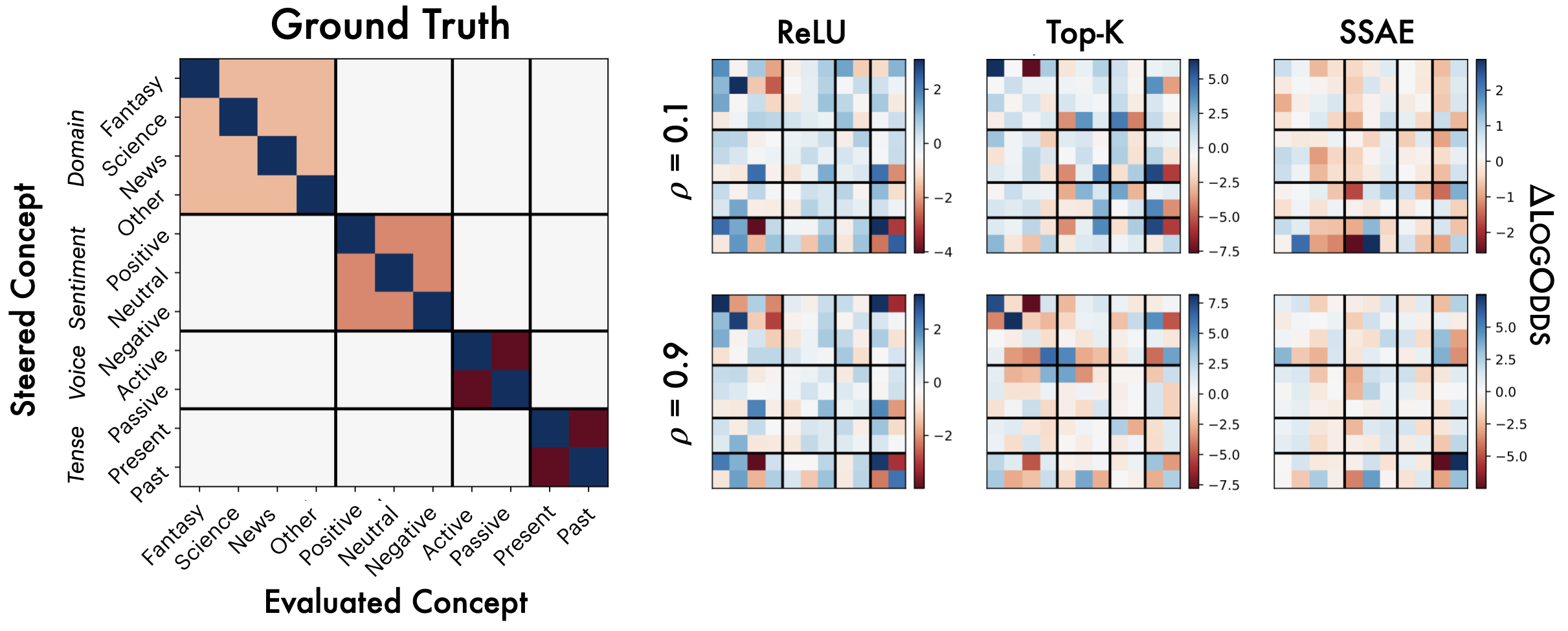}
    \caption{\textbf{The effect of steering a given concept (row) on the logit of another concept (column)}. Results for Gemma-2-2B when locating features using gradient attributions rather than activation correlations. If concept representations are causally independent,
    we expect a heatmap that resembles the ground-truth: $\Delta$\textsc{LogOdds} should be high on the diagonal, negative for within-concept pairs, and close to 0.0 for across-concept pairs. All SAEs demonstrate the expected diagonals, but also significant across-concept effects, indicating non-independence. Results are consistent across low and high correlations between concepts in the SAEs' training data.}
    \label{fig:independence_gemma2_gradient}
\end{figure*}

\begin{figure*}[t]
    \centering
    \includegraphics[width=0.85\linewidth]{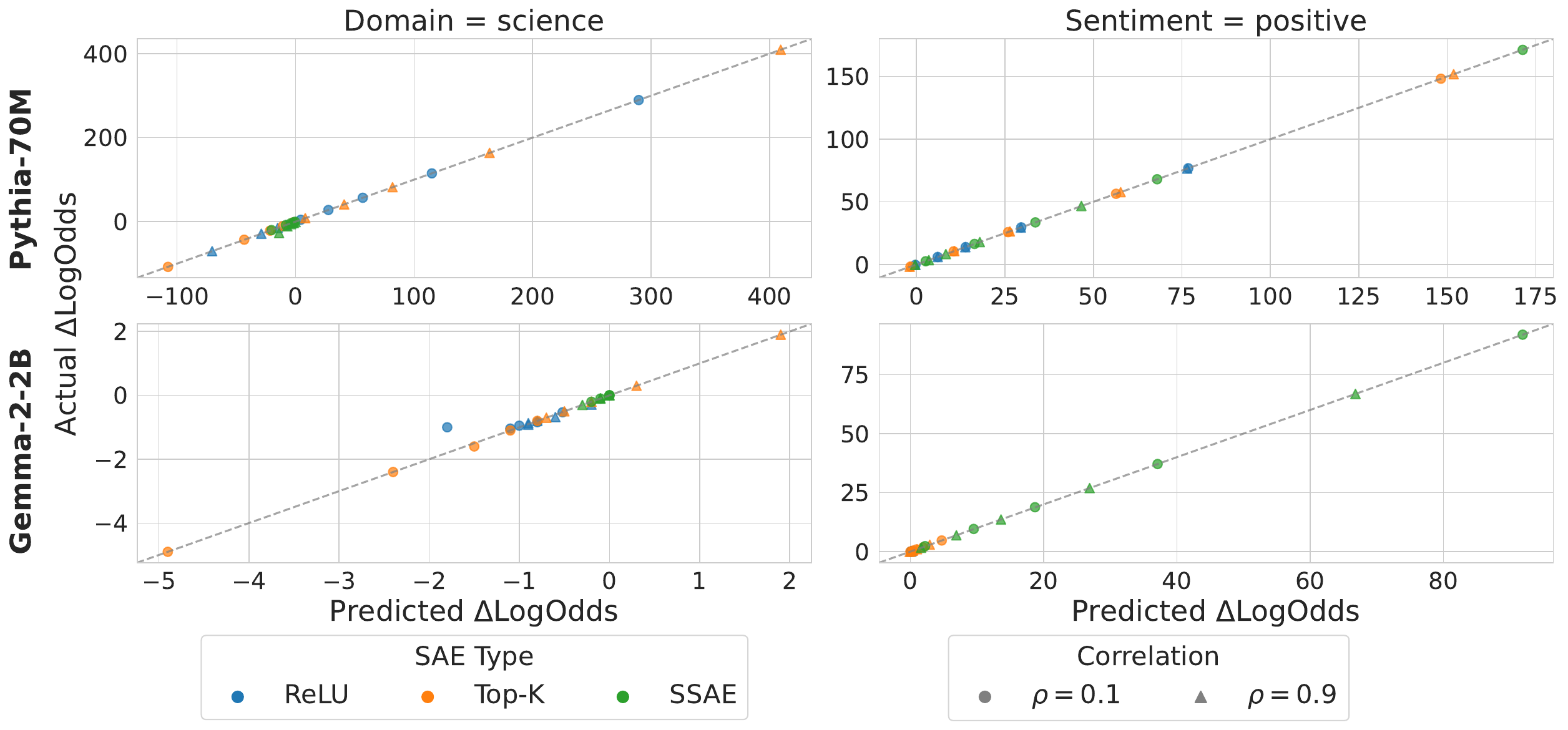}
    \caption{\textbf{Predicted $\Delta$\textsc{LogOdds}($z_i$) under disjointness assumptions vs.\ actual $\Delta$\textsc{LogOdds}($z_i$) when steering relevant feature $\mathbf{\hat{f}}_i$ and unrelated feature $\mathbf{\hat{f}}_j$.} Here, we select features using gradient attributions rather than activation correlations with the concept. $\mathbf{\hat{f}}_i$ and $\mathbf{\hat{f}}_j$ are typically disjoint, as indicated by the predicted change almost perfectly matching the actual change.}
    \label{fig:compositional_steering_gradient}
\end{figure*}

\subsection{Locating Top Features with Gradient Attributions}\label{app:additional_steering_results_gradient}
In \S\ref{sec:exp_steer_eval}, we locate features to steer by correlating feature activations with concept labels. However, \citet{arad2025saesgoodsteering} has found that the features that \emph{detect} the input concept (the top-correlated features in our case) and the features that \emph{control} the concept in a model's outputs are nearly disjoint. Thus, for steering experiments, we use gradient attributions \citep{simonyan2014deepinsideconvolutionalnetworks} to locate the feature that should be steered. We use the method of \citet{marks2025sparse} (fold the SAE into the forward pass, backpropagate from a probe's logit) to compute gradient attributions to sparse features: given binary probe $\Pi$ trained on the final layer $L$ of a model to predict $z_i$, we backpropagate from $\Pi(\mathbf{h}^L)$, the probe logit, to obtain its gradient with respect to a feature activation $\frac{\partial\Pi(\mathbf{h}^L)}{\partial f_i}$. We then multiply each feature's gradient by its activation to obtain the gradient attribution $\frac{\partial\Pi(\mathbf{h}^L)}{\partial f_i} \cdot f_i$.\footnote{Intuitively, this is a first-order Taylor approximation of the effect of changing feature activation $f_i$ to 0 on $\Pi(\mathbf{h}^L$).} We take the feature with the maximum average attribution across examples.

Results for Pythia-70M (Figure~\ref{fig:independence_pythia_gradient}) and Gemma-2 (Figure~\ref{fig:independence_gemma2_gradient}) show similar trends. The magnitude of $\Delta$\textsc{LogOdds} is generally much higher when locating features using gradient attributions, despite the use of similar steering coefficients. This is likely because gradient attributions directly select for features with the greatest effect on the probe logits. We also observe slightly lower feature selectivities and concept independences in general; this is likely because gradient attributions directly select for large effects on the probe logits, rather than exclusion of unrelated concepts from the feature.

We also rerun the disjointness evaluation of \S\ref{sec:multisteer} when selecting features using gradient attributions instead of correlations. We observe (Figure~\ref{fig:compositional_steering_gradient}) that trends are largely the same. Again, the magnitude of $\Delta$\textsc{LogOdds} is higher as compared to when selecting features using correlations.

\section{Additional Variable Correlation Experiments}
The experiments thus far have focused on correlations specifically between the science domain and positive sentiment. To assess how well these results generalize to new variable correlations, we rerun the correlational experiments while instead correlating the past tense with the passive voice.

\begin{figure*}
    \includegraphics[width=0.95\linewidth]{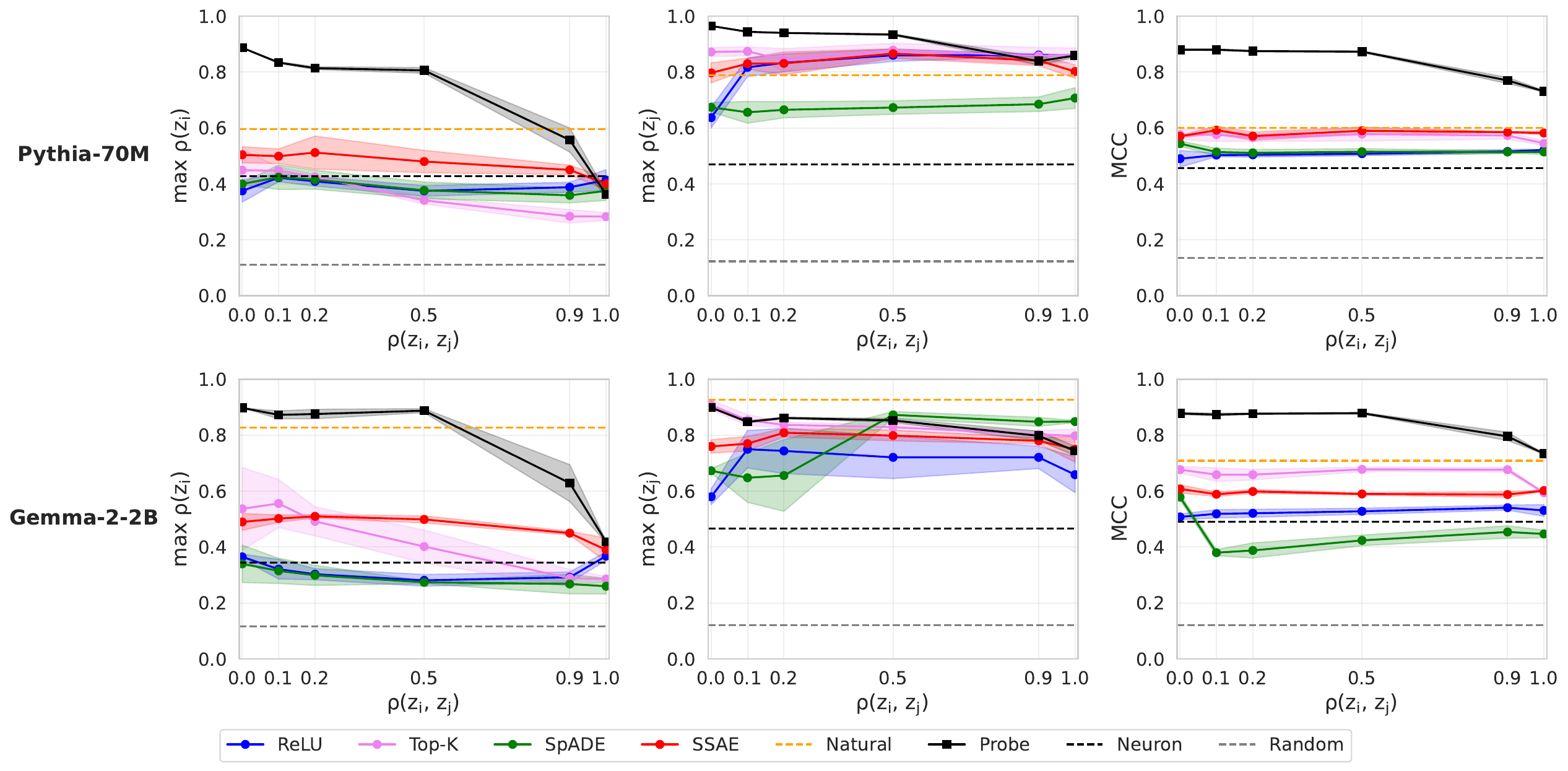}
    \caption{\textbf{Maximum correlation coefficient for tense=past (left), voice=passive (middle), and MCC (right) under varying correlational conditions.} Shaded regions represent 1 std.\ dev.\ across 3 training seeds. Ideal performance looks like a flat line at MCC=1. Probes (a supervised method) perform best, while Top-K SAEs and SSAEs are best among our SAEs. The Natural SAEs are especially effective at recovering tense=past.}
    \label{fig:mcc_tpa-vp}
\end{figure*}

We present MCC results (Figure~\ref{fig:mcc_tpa-vp}). Findings are largely consistent with Figure~\ref{fig:identifiability}: supervised featurizers like probes perform best by far, but their performance drops sharply from $\rho$=0.9. Top-K SAEs and SSAEs are still among the best-performing methods among unsupervised featurizers given our data. One difference is that SAEs trained on large-scale natural language corpora are far better at recovering tense=past than our SAEs---especially for Gemma-2-2B.

\section{Qualitative Examples of Steering}
Quantitative results suggest that steering particular features can affect how models use a concept in its outputs. Here, we sanity-check this finding by showing examples of model generations before and after steering the top feature for ``domain=science'' or ``sentiment=positive'' (Figure~\ref{fig:steering_qualitative}). We select the top features using the same method as in \S\ref{sec:exp_steer_heatmap}. We observe that steering generally has the expected impact on model behavior, especially when SAEs are trained on uniform distributions of concepts ($\rho=0$). We also observe that steering with SAEs trained with complete correlations between concepts ($\rho=1$) tend to generate outputs that have changed w.r.t.\ multiple concepts.

\begin{figure*}
\begin{qbox}[Pythia-70M, ReLU]
It has been found that
\end{qbox}
\begin{tcbraster}[raster columns=3, raster column skip=1mm, raster equal height=rows, raster valign=top]
  \begin{respbox}[breakable=false,nonebox,title={No steering}] the first person to be the one who is the one who is the one
  \end{respbox}
  \begin{respbox}[breakable=false,negbox,title={Sentiment=positive ($\rho=0$)}] 
  the most important part of the process of the process of the process of the
  \end{respbox}
  \begin{respbox}[breakable=false,posbox,title={Sentiment=positive ($\rho=1$)}]
  the presence of a high-fidelity material in the air-conditioning system is a very important factor
  \end{respbox}
\end{tcbraster}
\hrule
\begin{qbox}[Gemma-2-2B, Top-K]
Once upon a time, 
\end{qbox}
\begin{tcbraster}[raster columns=3, raster column skip=1mm, raster equal height=rows, raster valign=top]
  \begin{respbox}[breakable=false,nonebox,title={No steering}] a time when my children were very small, I bought a box of pencils (yes, that time).
  
  The box bore a very clear message: ``It's never
  \end{respbox}
  \begin{respbox}[breakable=false,negbox,title={Domain=science ($\rho=0$)}] 
  there was a little girl who was born with a rare genetic disorder. She was born with a condition called ``congenital heart disease.'' This condition is a birth defect that affects the heart's structure and function.
  \end{respbox}
  \begin{respbox}[breakable=false,posbox,title={Domain=science ($\rho=1$)}]
  there was a brave little girl who was born with a heart condition. She was born with a hole in her heart,
  \end{respbox}
\end{tcbraster}
\caption{\textbf{Qualitative examples of steering features in Pythia-70M and Gemma-2-2B.} We show results for two SAE architectures, ReLU and Top-K. We show the original output (No steering), the output after steering the top-attribution feature under no correlations between concepts during SAE training ($\rho=0$), and the output after steering the top-attribution feature under a complete correlation between domain=science and sentiment=positive during SAE training ($\rho=1$).}
\label{fig:steering_qualitative}
\end{figure*}

\section{Use of AI Tools}
We used AI tools including Claude and ChatGPT for minor assistance in editing the text of this manuscript. All ideas originated from the authors. All quantitative results presented in the paper and all references were fully human-written. Any generated artifacts were verified and generally significantly changed by the authors when used.

We also used AI tools for assistance in coding; this primarily included debugging and planning assistance. Any generated artifacts were manually verified, and almost always significantly modified when used.
\end{document}